\documentclass[10pt]{article}

\PassOptionsToPackage{table}{xcolor}
\usepackage{paperstyle}                  

\usepackage{lipsum}                 
\usepackage{multirow}                                                                                                                                                                                                                                  
\usepackage{latexsym}
\usepackage{algorithm}  
\usepackage{eso-pic}
\usepackage{graphicx}                                                                                                        \usepackage[utf8]{inputenc}                                                       
\usepackage{algpseudocode}
  \usepackage{wrapfig}
  \usepackage{url}
  \usepackage{booktabs}
  \usepackage{amsfonts}
  \usepackage{nicefrac}
  \usepackage{textcomp}

  \renewcommand{\arraystretch}{1.15}
  \definecolor{ModelGreen}{RGB}{215,230,212}
\usepackage{adjustbox} 
\usepackage{float}

                                                            \AddToShipoutPictureFG*{
    \put(90,735){
        \includegraphics[width=0.7\paperwidth]{logo_banner_hires.png}
    }
}                                                                                      \DeclareUnicodeCharacter{22C6}{\textasteriskcentered}

\title{DyCo-RL: Dynamic Cross-Modal Coordination for Visual Reasoning}

\author{Hangui Lin$^{2\dagger}$, Yan Shu$^{1\dagger}$, Chi Liu$^{3}$, Zhengyang Liang$^{4}$, Xiangrui Liu$^{2}$, Minghao Qin$^{2}$, Derek Li$^{3}$, Bryan Dai$^{3}$, Teng Long$^{1}$, Zheng Liu$^{2}$, Nicu Sebe$^{1}$\\[3pt]
{\normalfont $^{1}$University of Trento\quad$^{2}$BAAI\quad$^{3}$IQuest Research\quad$^{4}$Singapore Management University}\\[2pt]
}

\begin{document}

\maketitle
\begin{abstract}
Reinforcement Learning with Verifiable Rewards (RLVR) has emerged as a leading paradigm for enhancing visual reasoning in Multimodal Large Language Models (MLLMs). However, existing RLVR methods optimize primarily for the reasoning outcome and overlook how that outcome is produced: as a chain-of-thought unfolds, some tokens must extract evidence from the image while others must anchor to the preceding textual context, yet a single sequence-level reward provides no signal about whether each token draws on the modality its role demands. Token-level analyses and controlled interventions reveal that tokens frequently under-attend to their designated modality,
and that this coordination breakdown may contribute to reasoning failures. Motivated by these findings, we propose DyCo-RL, which integrates dynamic cross-modal coordination into RLVR optimization. Specifically, DyCo-RL uses the Fisher–Rao geodesic distance to measure within-modality attention shifts, assigning tokens to either visually-oriented or text-oriented functional roles. It then evaluates the alignment between a token's actual attention allocation and its assigned role, leveraging this score for alignment-guided advantage reweighting during policy optimization. Extensive experiments using 11K examples for RL training demonstrate that the algorithm-agnostic DyCo-RL, when applied to Qwen2.5-VL-3B/7B and Qwen3.5-27B,
improves the average performance of all four evaluated RLVR algorithms across seven benchmarks spanning visual-centric and mathematical reasoning. Our repository is available at \url{https://github.com/Sammy20207109/DyCo-RL}
\\

{\fontsize{10pt}{12pt}\bfseries\selectfont
$\dagger$ Equal Contribution
}
\end{abstract}
\vspace{-10pt}
\section{Introduction}
\vspace{-5pt}

Visual reasoning~\citep{ke2025explainanswersurveycompositional, HE2021104194} requires progressively acquiring, interpreting, and integrating information across visual and textual modalities. Reinforcement Learning with Verifiable Rewards (RLVR)~\citep{shao2024deepseekmathpushinglimitsmathematical,zheng2025groupsequencepolicyoptimization,gao2025softadaptivepolicyoptimization,yu2025dapoopensourcellmreinforcement}, originally developed for LLMs, has recently been extended to MLLMs as a unified framework that jointly improves visual perception and multi-step reasoning through end-to-end reward-driven training~\citep{wang2025perception, huang2025spotlight,rezaei2026cppo,zhang2025perceptual}.

\begin{figure*}[t]
    \centering
    \includegraphics[width=1\linewidth]{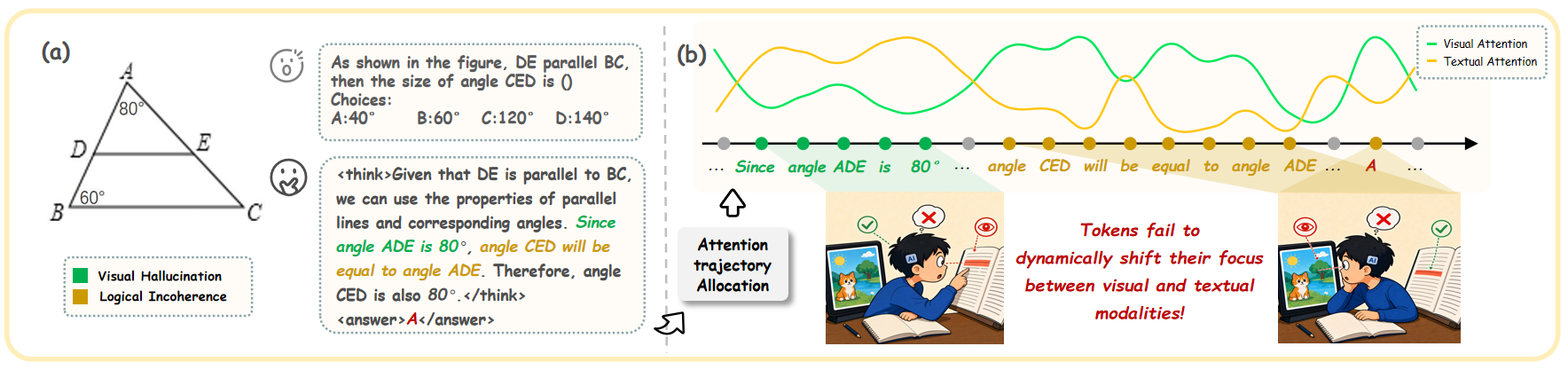}
   \vspace{-15pt}
    \caption{An illustrative example of reasoning failures in Qwen2.5-VL-3B~\citep{bai2025qwen25vltechnicalreport} optimized via GRPO on ThinkLite-hard-11K~\citep{wang2025sota}. (\textbf{a}) The model first hallucinates visual features (mistakenly asserting $\angle \text{ADE} = 80^{\circ}$), and subsequently generates logically incoherent text (baselessly claiming $\angle \text{CED} = \angle \text{ADE}$, contradicting its own reasoning history). (\textbf{b}) Token-level attention trajectory analysis reveals the coordination breakdown behind these errors: visually-oriented tokens under-attend to the image, while text-oriented tokens fail to sufficiently anchor to the preceding textual context.}
    \label{fig:1}
    \vspace{-5pt}
\end{figure*}

However, visual reasoning is inherently a dynamic process. As the Chain-of-Thought (CoT) unfolds, the model must continuously alternate between faithfully extracting visual evidence at certain steps and strictly anchoring its logic to the preceding textual context at others. When this cross-modal coordination breaks down, distinct failure modes emerge. As illustrated in Fig.~\ref{fig:1}(a), the model first hallucinates visual features by mistakenly asserting that angle ADE is 80°, and subsequently suffers from logical incoherence by baselessly claiming that angle CED equals angle ADE without valid justification from its own reasoning history. Existing RLVR methods, however, provide limited explicit guidance for coordinating visual evidence extraction and textual-context anchoring according to each token's functional role.
They either broadcast a single sequence-level advantage uniformly to all tokens~\citep{shao2024deepseekmathpushinglimitsmathematical,zheng2025groupsequencepolicyoptimization}, or reinforce visual perception alone~\citep{lookback2025,wang2025perception,huang2025spotlight}. In both cases, whether a token draws on the modality its current step requires is never supervised.

To uncover the mechanism behind these failures, we examine the rationale generation process at the level of token attention, tracking how each token distributes its attention between visual patches and the textual context~\citep{zhang2025mllms,HASSANIN2024102417,10897531}. As illustrated in Fig.~\ref{fig:1}(b), we first observe a \textit{correlation}: in erroneous samples, visually-oriented tokens systematically under-attend to the image, while text-oriented tokens fail to sufficiently anchor to the preceding textual context.
Crucially, we further provide \textit{causal evidence} through controlled intervention experiments. We demonstrate that dynamically amplifying the appropriate modality attention based on a token's functional role can rectify reasoning failures and recover correct answers.

Building on these findings, we propose \textbf{Dy}namic \textbf{Co}ordination \textbf{R}einforcement \textbf{L}earning (\textbf{DyCo-RL}) to integrate token-level cross-modal coordination directly into the RLVR framework. DyCo-RL operates in two stages. 
First, by computing the Fisher--Rao geodesic distance between consecutive within-modality attention distributions, it assigns each token a functional role
(i.e., visually-oriented, text-oriented, or neutral); a larger distance reflects greater attention redistribution, which we use as a proxy for active information extraction from that modality.
Second, it evaluates the alignment between the token's actual attention allocation and its assigned role. This alignment score is then leveraged for alignment-guided advantage reweighting during policy optimization, effectively amplifying the learning signals for well-coordinated tokens while attenuating the influence of misaligned ones. Designed as an algorithm-agnostic plug-in, DyCo-RL can be seamlessly incorporated into mainstream RL algorithms. In summary, our main contributions are as follows:

\begin{itemize}
    \item We identify cross-modal coordination breakdowns as a critical bottleneck in visual reasoning for MLLMs. Through token-level correlational analysis and causal intervention experiments, we reveal that tokens frequently fail to dynamically shift their focus between modalities during CoT generation.
    
    \item We propose DyCo-RL, a novel approach that embeds dynamic, token-level cross-modal coordination into the RLVR framework. By leveraging the Fisher--Rao geodesic distance of attention shifts to assign functional roles, DyCo-RL dynamically reweights token-level advantages based on modality-specific alignment signals.


    \item Using only 11K training examples, DyCo-RL improves seven-benchmark
average scores across Qwen2.5-VL-3B/7B and Qwen3.5-27B with GRPO,
DAPO, SAPO, and GSPO, gaining up to 3.2 points.
Additional comparisons show advantages over related methods,
while ablations support role-conditioned scoring.
\end{itemize}

\vspace{-6pt}
\section{Related Work}
\vspace{-5pt}
\textbf{RLVR for Visual Reasoning.}
Reinforcement Learning with Verifiable Rewards (RLVR) has emerged as a dominant paradigm for enhancing visual reasoning in MLLMs. Foundational algorithms such as GRPO~\citep{shao2024deepseekmathpushinglimitsmathematical}, DAPO~\citep{yu2025dapoopensourcellmreinforcement}, GSPO~\citep{zheng2025groupsequencepolicyoptimization}, and SAPO~\citep{gao2025softadaptivepolicyoptimization} derive advantages from sequence-level rewards, applying uniform credit assignment across tokens. To explicitly enhance visual perception, PAPO~\citep{wang2025perception} introduces an implicit perception loss via KL divergence between outputs under original and masked visual inputs, encouraging visually grounded reasoning at the loss level. Progressing toward finer granularity,
StepGRPO~\citep{zhang2025r1vl} incorporates dense step-level rewards to evaluate the accuracy and logical consistency of intermediate reasoning steps. Finally, at the token level, fine-grained optimization has been explored for preference alignment in LLMs~\citep{zeng2024token,pmlr-v267-lin25j} and entropy-based token selection in RLVR~\citep{wang2025beyond8020}. In the multimodal domain, VPPO~\citep{huang2025spotlight} filters policy gradients to emphasize highly visually-dependent tokens, and AT-RL~\citep{jiao2026creditduecrossmodalityconnectivity} selectively reinforces anchor tokens exhibiting strong cross-modal connectivity via graph-based attention clustering.
Despite these advances, explicit guidance for credit assignment that jointly accounts for visual evidence extraction and reasoning over the textual context remains limited. Our work addresses this gap by reweighting token-level advantages according to role-specific modality alignment.

\textbf{Cross-Modal Attention Analysis in MLLMs.}
Attention patterns serve as a crucial window into the cross-modal interactions of MLLMs.
One line of research explores inference-time interventions, including attention amplification, attention-guided visual cropping, and visual-aware decoding~\citep{liu2024pai,zhang2025mllms,ICLR2025_24079b91}, to improve visual perception and mitigate hallucinations. Another line of research investigates attention from an interpretability perspective.
Recent studies show that models can attend to relevant image regions even when producing incorrect answers~\citep{zhang2025mllms}, and that visual attention may decline during extended reasoning~\citep{sun2025tvc}, motivating closer examination of attention allocation throughout generation.
Attention has also been incorporated into RL through reward design and token selection~\citep{jian2025reflectionv,jiao2026creditduecrossmodalityconnectivity}.
Building on these insights, DyCo-RL uses within-modality attention shifts as a proxy for functional role assignment, translating attention dynamics into role-conditioned guidance for RLVR optimization.

\vspace{-3pt}
\section{Preliminaries}
\vspace{-3pt}
\subsection{Modality-Specific Attention Decomposition}
\vspace{-2pt}

In autoregressive MLLMs, the decoder attends to both visual and
textual tokens at each generation step. Let $\mathcal{I}_{\text{vis}}$
denote the indices of visual tokens and $\mathcal{I}_{\text{txt}}$
the indices of non-system textual tokens, including the user
prompt, associated special tokens, and all  generated
tokens.  
At the last decoder layer, the attention weight at decoding step $t$ to a
context position $j \le t$, averaged over heads, is
\begin{equation}
\small
a_{t,j} = \frac{1}{|\mathcal{H}|} \sum_{h \in \mathcal{H}}
\frac{\exp(e^{h}_{tj})}{\sum_{k \le t} \exp(e^{h}_{tk})}, \qquad
e^{h}_{tj} =
\frac{(\mathbf{q}^{h}_{t})^{\top}\mathbf{k}^{h}_{j}}{\sqrt{d}},
\end{equation}
where $\mathcal{H}$ denotes the set of attention heads,
$\mathbf{q}^{h}_{t}$ and $\mathbf{k}^{h}_{j}$ are the query and key
vectors of head $h$, respectively, and $d$ is the head dimension. The \emph{modality-specific attention mass} of step $t$ on modality $m \in \{\text{vis}, \text{txt}\}$ is
\vspace{-1pt}
\begin{equation}
\small
r^{m}_{t} = \sum_{j \in \mathcal{I}_m} a_{t,j},
\end{equation}
\vspace{-5pt}
which measures how much attention the generated token allocates to modality $m$ in total. 
\vspace{-5pt}
\subsection{Group Relative Policy Optimization (GRPO)}
\vspace{-5pt}
Given an input prompt $x$, the current policy $\pi_{\theta_{\text{old}}}$ (i.e., the policy before the update) samples a group of $G$ candidate responses $\{o_1, \dots, o_G\}$. Under the RLVR framework, each response receives an outcome-based reward $R_i$ determined by whether its final answer matches the ground truth. To reduce variance without a separate value network, GRPO computes a group-normalized advantage for each response:
\vspace{-1pt}
\begin{equation}
\small
\hat{A}_i = \frac{R_i - \mu(R_{1:G})}{\sigma(R_{1:G})},
\end{equation}
where $\mu(R_{1:G})$ and $\sigma(R_{1:G})$ are the mean and standard deviation of the rewards within the group. The policy $\pi_\theta$ is then optimized by maximizing a clipped surrogate objective, in which the scalar advantage $\hat{A}_i$ is applied to every token position of the response:
\begin{equation}
\small
\label{eq:grpo}
\mathcal{J}_{\text{GRPO}}(\theta) = \mathbb{E}_{x \sim \mathcal{X},\ \{o_i\}_{i=1}^{G} \sim \pi_{\theta_{\text{old}}}(\cdot \mid x)}\Bigg[ \frac{1}{G}\sum_{i=1}^G \frac{1}{|o_i|}\sum_{t=1}^{|o_i|} \min\Big(\rho_{i,t}\,\hat{A}_i,\ 
\text{clip}\big(\rho_{i,t}, 1{-}\epsilon, 1{+}\epsilon\big)\hat{A}_i\Big) \Bigg],
\end{equation}

where $\rho_{i,t} = \pi_\theta(o_{i,t} \mid x, o_{i,<t})\,/\,\pi_{\theta_{\text{old}}}(o_{i,t} \mid x, o_{i,<t})$ is the importance sampling ratio of the $t$-th token, and the KL regularization term is omitted for brevity.

Since $\hat{A}_i$ is shared by all tokens of $o_i$, a token that reads a value from the image and a token that derives a conclusion from the preceding text receive the same learning signal, regardless of whether each attends to the modality its step requires.

\vspace{-10pt}
\section{Method}
\vspace{-8pt}
We first analyze the cause of cross-modal coordination breakdowns in visual reasoning (Section~\ref{4.1}). Building on these findings, we propose \textbf{Dy}namic \textbf{Co}ordination \textbf{R}einforcement \textbf{L}earning (\textbf{DyCo-RL}), which operates in two stages: (i) assigning tokens to visually- or text-oriented  roles via the Fisher--Rao geodesic distance between consecutive within-modality attention distributions (Section~\ref{4.2.1}); and (ii) evaluating the alignment between actual attention allocation and assigned roles to perform alignment-guided advantage reweighting during policy optimization (Section~\ref{4.2.2}).

\vspace{-8pt}
\subsection{Diagnosing Cross-Modal Coordination Breakdowns}
\label{4.1}
\vspace{-8pt}
We examine cross-modal coordination breakdowns at token-level granularity from two complementary perspectives: \textbf{correlation analysis}, which examines whether attention patterns systematically differ between correct and erroneous tokens, and \textbf{causal intervention}, which tests whether correcting attention misalignment recovers model performance.
\vspace{-10pt}
\paragraph{Correlation Analysis.}

\begin{figure}[t]
\vspace{-5pt}
    \centering
    \includegraphics[width=0.8\linewidth]{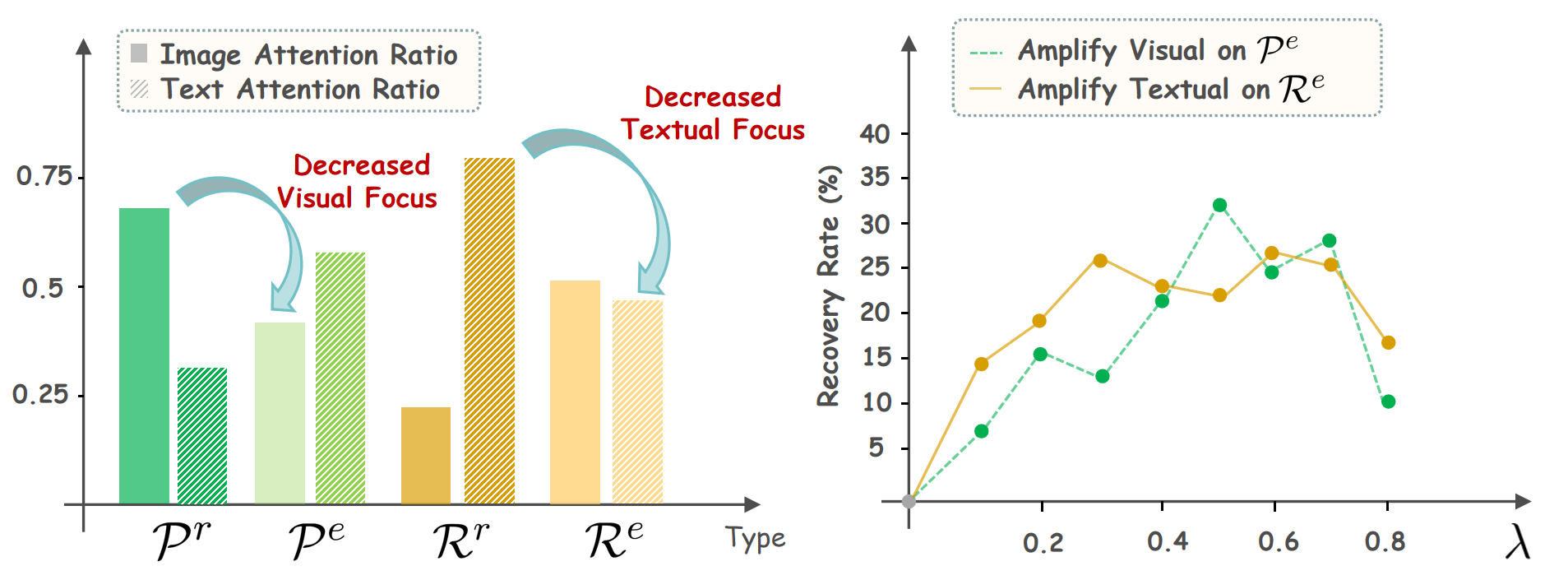}
    \vspace{-5pt}
   \caption{\textbf{(Left)} Correlation analysis of modality-specific attention. $\mathcal{P}$/$\mathcal{R}$ denote visually/text-oriented tokens, and superscripts $r$/$e$ indicate correct/erroneous states. Attention ratios are computed as
 $\small r_t^m/(r_t^{\mathrm{vis}}+r_t^{\mathrm{txt}})$. \textbf{(Right)} Causal intervention on erroneous tokens. The green (yellow) curve shows the effect of amplifying visual (textual) attention for $\mathcal{P}^e$ ($\mathcal{R}^e$). Both targeted interventions yield a consistent recovery rate under moderate enhancement.}
    \label{fig:relevant}
    \vspace{-5pt}
\end{figure}

We collect 200 erroneous rollouts from MathVerse~\citep{zhang2024mathverse} and MathVision~\citep{wang2024measuring} generated by Qwen2.5-VL-3B optimized via GRPO on ThinkLite-hard-11K. Each generated token span is categorized according to whether it primarily performs image-grounded perception or text-based reasoning within the local decoding context. Specifically, we distinguish \textit{visually-oriented} tokens ($\mathcal{P}$), which extract or describe information from the image, from \textit{text-oriented} tokens ($\mathcal{R}$), which conduct logical inference based on the preceding textual context. 
Each category is further divided by correctness, yielding four disjoint groups: $\mathcal{P}^r$ (correct), $\mathcal{P}^e$ (erroneous), $\mathcal{R}^r$ (correct), and $\mathcal{R}^e$ (erroneous). Labels are assigned by annotators at the semantic-span level and then projected back to tokens. Full annotation details are provided in Appendix~\ref{app:hal_analysis}.

For each group $\mathcal{S} \in \{\mathcal{P}^r, \mathcal{P}^e, \mathcal{R}^r, \mathcal{R}^e\}$,
we average the modality-specific attention ratio $r_t^m/(r_t^{\mathrm{vis}}+r_t^{\mathrm{txt}})$ over all tokens in the group. As shown in Fig.~\ref{fig:relevant}, correct visually-oriented tokens ($\mathcal{P}^r$) allocate significantly more attention to the image than erroneous ones
($\mathcal{P}^e$). Symmetrically, correct text-oriented tokens ($\mathcal{R}^r$) attend more to the preceding context than erroneous ones ($\mathcal{R}^e$). These results reveal a clear association between modality-specific attention allocation and token-level correctness. Additional correlation analysis is provided in Appendix~\ref{app:position_analysis}.  
\vspace{-10pt}
\paragraph{Causal Intervention.}
To test whether the observed attention misalignment causally contributes to reasoning errors, rather than merely correlating with them, we design a controlled attention enhancement experiment. For tokens in $\mathcal{P}^e$ or $\mathcal{R}^e$, we selectively amplify their attention toward the modality their functional role demands: visual inputs for visually-oriented erroneous tokens ($\mathcal{M} = \mathcal{I}_{\text{vis}}$) and textual context for text-oriented erroneous tokens ($\mathcal{M} = \mathcal{I}_{\text{txt}}$). Specifically, at each intervened decoding step $t$, we enhance attention toward the target modality independently within each head of the last decoder layer:
\vspace{-3pt}
\begin{equation}
\label{eq:attention_intervention}
\small
\tilde{a}^{h}_{t,j}
=
\frac{
\exp(e^{h}_{tj})
\bigl(1+\lambda\mathbf{1}_{j\in\mathcal{M}}\bigr)
}{
\sum_{k\le t}
\exp(e^{h}_{tk})
\bigl(1+\lambda\mathbf{1}_{k\in\mathcal{M}}\bigr)
}.
\end{equation}
Here, $e^{h}_{tj}$ is the attention logit defined in Eq.~(1),
$\tilde{a}^{h}_{t,j}$ is the modified attention weight of head $h$,
and $\lambda>0$ controls the enhancement strength.
The modified weights are normalized separately within each head.
Details of the replay protocol and control conditions are provided
in Appendix~\ref{app:intervention_controls}.

As shown in Fig.~\ref{fig:relevant},
moderate enhancement can recover correct answers on previously erroneous samples for both functional roles, with a higher peak recovery rate for visually-oriented interventions.
Excessively large $\lambda$, however, degrades performance, indicating that over-amplification disrupts the original attention balance. The role-aware intervention also outperforms the controls at most tested strengths (Appendix~\ref{app:intervention_controls}).
Together with the correlation analysis, these findings provide causal evidence that role-aligned attention modulation can help correct visual reasoning errors.


\vspace{-3pt}
\subsection{Dynamic Coordination Reinforcement Learning}
\begin{figure*}[t]
    \centering
    \includegraphics[width=1\linewidth]{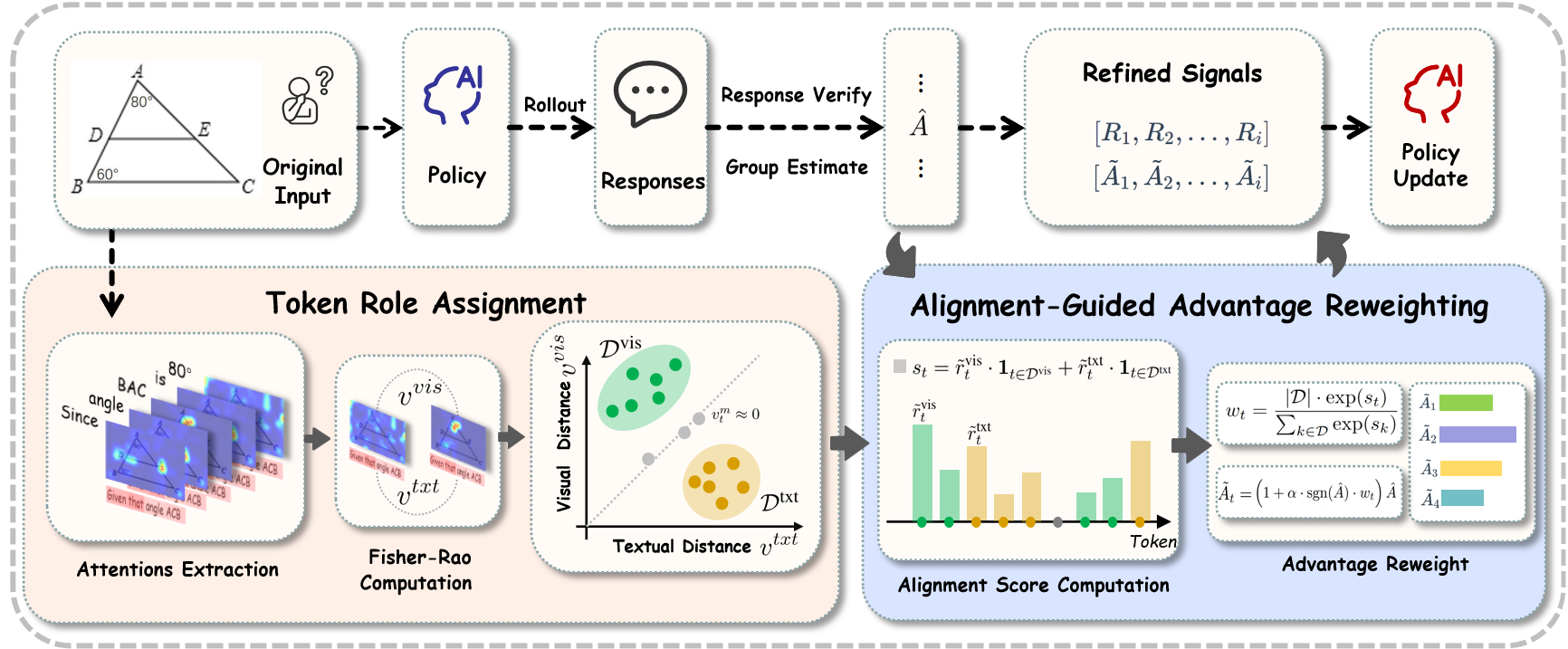}
    \vspace{-15pt}
   \caption{\textbf{Overall Pipeline of DyCo-RL.}  Instead of broadcasting the sequence-level advantage $\hat{A}_i$ to all tokens, DyCo-RL assigns each token a functional role via the Fisher--Rao distance between consecutive within-modality attention distributions, scores how well the token's attention matches that role, and uses the score to reweight $\hat{A}_i$ at the token level, yielding $\tilde{A}_{i,t}$ for the policy update.}
    \label{fig:pipline}
    \vspace{-10pt}
\end{figure*}

\vspace{-3pt}
Based on the findings above, we propose DyCo-RL, whose pipeline is illustrated in Fig.~\ref{fig:pipline}.

\vspace{-5pt}
\subsubsection{Token Role Assignment}
\label{4.2.1}
\vspace{-5pt}

Acquiring new information from a modality may involve redistributing attention across positions within that modality. We therefore characterize a token's functional role through changes in its within-modality attention distribution between consecutive steps. Total attention mass measures how much attention a modality receives, but does not distinguish persistent attention patterns, such as attention sinks~\citep{queipo2026attention}, from step-specific shifts in focus. We use within-modality redistribution to infer functional roles and, once a role is assigned, use modality-level attention mass to score the attention allocated to the corresponding modality.

First, we normalize the attention weights within each modality to obtain a conditional distribution:

\vspace{-7pt}
\begin{equation}
\small
p^{m}_{t,j} = \frac{a_{t,j}}{r^{m}_t}, \quad \forall j \in \mathcal{I}_m
\end{equation}
\vspace{-1pt}
which describes how attention is distributed across context tokens within modality $m$, independent of the total amount allocated to that modality.

Next, we measure the structural change between consecutive within-modality distributions using the Fisher--Rao geodesic distance:
\vspace{-7pt}
\begin{equation}
\small
v^{m}_{t} = 2\arccos\!\bigg(\sum_{j \in \mathcal{I}_m} \sqrt{p^{m}_{t-1,\,j}\; p^{m}_{t,\,j}}\bigg)
\end{equation}
\vspace{-1pt}
For the textual modality, we restrict both consecutive distributions to their shared prefix and independently renormalize them before computing the distance.
A high $v^{m}_t$ indicates substantial within-modality attention redistribution, which we use as a proxy for active information acquisition, whereas $v^{m}_t \approx 0$ indicates little change in the attention distribution.
Under the square-root embedding $p\mapsto 2\sqrt{p}$, the Fisher--Rao distance measures spherical geodesic separation between attention distributions, providing a symmetric measure bounded in $[0,\pi]$.

Finally, we assign each token to the modality whose attention distribution undergoes stronger structural reconfiguration:
\vspace{-2pt}
\begin{equation}
\small
\mathcal{D}^{m} = \big\{\, t \mid v^{m}_{t} - v^{\bar{m}}_{t} > \tau \,\big\}, \quad \{m, \bar{m}\} = \{\text{vis}, \text{txt}\}
\end{equation}
\vspace{-2pt}
where $\tau$ is a stability margin that suppresses ambiguous transitions, and $\bar{m}$ denotes the complementary modality. Tokens falling outside both sets ($\mathcal{D}^{\text{vis}}$ and $\mathcal{D}^{\text{txt}}$) are treated as neutral tokens where neither modality exhibits dominant behavior.

\subsubsection{Alignment-Guided Advantage Reweighting}
\label{4.2.2}
Building on the proposed token role assignment, we translate modality
alignment into fine-grained RL optimization guidance.
To derive token-level operations, we temporarily drop the response
index $i$ and denote the response length by $T$.
For each modality $m$, we min--max normalize its attention
mass across the response:
$\tilde{r}_t^m =
\frac{r_t^m-\min_k r_k^m}{\max_k r_k^m-\min_k r_k^m}$,
where $t,k\in\{1,\ldots,T\}$.
The alignment score selects the normalized attention mass
corresponding to the token's assigned functional role:
\begin{equation}
\small
\label{eq_attention}
s_t = \tilde{r}_t^{\text{vis}} \cdot \mathbf{1}_{t \in \mathcal{D}^{\text{vis}}} + \tilde{r}_t^{\text{txt}} \cdot \mathbf{1}_{t \in \mathcal{D}^{\text{txt}}}.
\end{equation}
Let $\mathcal{D}=\mathcal{D}^{\mathrm{vis}}\cup\mathcal{D}^{\mathrm{txt}}$ denote the set of non-neutral tokens. Tokens in $\mathcal{D}$ have their advantage reweighted by the alignment score, while neutral tokens retain the sequence-level advantage:
\begin{equation}
\tilde{A}_{t} =
\left\{
\begin{array}{cl}
\left(1+\alpha \cdot \operatorname{sgn}(\hat{A}) \cdot w_t\right)\hat{A},
& t\in\mathcal{D},\\[2pt]
\hat{A},
& t\notin\mathcal{D},
\end{array}
\right.
\quad
w_t =
\frac{|\mathcal{D}| \cdot \exp(s_t)}
{\sum_{k\in\mathcal{D}} \exp(s_k)},
\quad t\in\mathcal{D}.
\end{equation}
where $\alpha > 0$ controls the reweighting strength and $|\mathcal{D}|$ normalizes $w_t$ to unit mean over $\mathcal{D}$; a centered variant removing the resulting common shift is evaluated in Table~\ref{tab:centered_reweighting}.

The reweighting thus reinforces role-aligned behavior by
amplifying positive advantages and attenuating negative ones
more strongly for tokens with higher alignment scores,
encouraging the model to read the image or use the preceding
context when the role requires it.
Substituting $\tilde{A}_{i,t}$ for $\hat{A}_i$ in Eq.~\eqref{eq:grpo}, the DyCo-RL objective with GRPO as the base algorithm is
\begin{equation}
\small 
\mathcal{J}_{\text{DyCo-RL}}(\theta) = \mathbb{E}_{x \sim \mathcal{X},\ \{o_i\}_{i=1}^{G} \sim \pi_{\theta_{\text{old}}}(\cdot \mid x)}\Bigg[\frac{1}{G}\sum_{i=1}^{G}\frac{1}{|o_i|}\sum_{t=1}^{|o_i|} \min\Big(\rho_{i,t}\,\tilde{A}_{i,t},\ 
\text{clip}\big(\rho_{i,t}, 1{-}\epsilon, 1{+}\epsilon\big)\,\tilde{A}_{i,t}\Big)\Bigg].
\end{equation}

\vspace{-10pt}
\section{Experiments}
\label{sec:experiments}
\vspace{-8pt}

\subsection{Experimental Settings}
\vspace{-5pt}
\paragraph{Models, Data, and Baselines.}
We integrate DyCo-RL into Qwen2.5-VL-3B/7B and Qwen3.5-27B, utilizing the ThinkLite-hard-11K dataset for training. Qwen2.5-VL is a commonly used open-source backbone in recent MLLM reinforcement-learning studies, enabling comparisons under a stable architecture and helping isolate the effect of the optimization method. We further include Qwen3.5-27B~\citep{qwen3.5} as a newer and substantially larger backbone to examine whether the benefits of DyCo-RL generalize across model generations and scales, rather than being specific to Qwen2.5-VL. ThinkLite-hard-11K provides a compact yet challenging set of difficulty-curated multimodal reasoning samples selected via MCTS from a substantially larger data pool, making it well suited for resource-efficient RLVR training. To demonstrate its versatility, we evaluate DyCo-RL on top of four representative RLVR algorithms spanning diverse optimization paradigms: GRPO~\citep{shao2024deepseekmathpushinglimitsmathematical}, DAPO~\citep{yu2025dapoopensourcellmreinforcement}, GSPO~\citep{zheng2025groupsequencepolicyoptimization}, and SAPO~\citep{gao2025softadaptivepolicyoptimization}. Brief descriptions of each baseline algorithm are provided in Appendix~\ref{app:alg}.
\vspace{-10pt}
\paragraph{Training Details.}
Due to limited computational resources, we train on the 11K samples for one epoch, with the main results reported using random seed 42. Multi-seed results are provided in Appendix~\ref{app:multiseed}. Models are trained with a learning rate of 1e-6, 4 rollouts per prompt, a maximum response length of 2048 tokens, and a global batch size of 64. For our DyCo-RL plugin, we set the alignment-guided reweighting strength $\alpha = 0.2$ and the token role assignment stability margin $\tau = 0.05$. Algorithm-specific hyperparameters of all RLVR baselines follow their official implementations. Comprehensive training configurations are detailed in Appendix~\ref{appendix:settings}.
\vspace{-2pt}
\paragraph{Evaluation Benchmarks.}
We evaluate on seven benchmarks spanning two primary domains: (1) \textit{Mathematical Reasoning}: WeMath~\citep{qiao2025we}, MathVision~\citep{wang2024measuring}, and MathVerse~\citep{zhang2024mathverse}; and (2) \textit{Visual-Centric Reasoning}: LogicVista~\citep{xiao2024logicvista}, HallusionBench~\citep{Guan_2024_CVPR}, MME~\citep{fu2023mme}, and MMBench~\citep{liu2024mmbench}. 
Comprehensive evaluation configurations are detailed in Appendix~\ref{app:alg}.
\vspace{-7pt}
\subsection{Main Results}
\vspace{-5pt}
\label{sec:main_results}
\begin{table*}[t]
\vspace{-12pt}
\scriptsize
\centering
\caption{Main Results across four RLVR algorithms and three model scales. Bold formatting indicates the better performance between each baseline algorithm and its DyCo-RL enhanced counterpart.}

\label{tab:main}

\setlength{\tabcolsep}{5pt}
\renewcommand{\arraystretch}{1.15}

\begin{adjustbox}{max width=\linewidth}
\begin{tabular}{lcccccccc}
\toprule

\multirow{2}{*}{\textbf{Model}}
& \multicolumn{3}{c}{\textbf{Mathematical Reasoning}}
& \multicolumn{5}{c}{\textbf{Visual-Centric Reasoning}} \\

\cmidrule(lr){2-4}
\cmidrule(lr){5-8}

& \textbf{WeMath}
& \textbf{MathVision}
& \textbf{MathVerse}
& \textbf{MMBench}
& \textbf{LogicVista}
& \textbf{HallusionBench}
& \textbf{MME}
& \textbf{Avg.} \\

\midrule

\rowcolor{gray!15}
\multicolumn{9}{c}{\textbf{Qwen2.5-VL-3B}} \\

\quad \textit{Zero-Shot}
& 20.0
& 23.0
& 33.8
& 71.4
& 34.5
& 61.3
& 76.9
& 45.8 \\

\quad GRPO
& 23.0
& 21.4
& 34.4
& 72.8
& 39.8
& 60.8
& 79.9
& 47.4 \\

\rowcolor{ModelGreen}
\quad \textbf{GRPO + DyCo-RL}
& \textbf{25.9}
& \textbf{22.4}
& \textbf{36.0}
& \textbf{76.2}
& \textbf{41.2}
& \textbf{62.4}
& \textbf{80.7}
& \textbf{49.3} \\

\quad DAPO
& \textbf{27.4}
& 17.1
& 34.0
& 73.8
& 39.8
& 60.7
& 81.0
& 47.7 \\

\rowcolor{ModelGreen}
\quad \textbf{DAPO + DyCo-RL}
& 26.9
& \textbf{23.0}
& \textbf{36.6}
& \textbf{75.6}
& \textbf{41.4}
& \textbf{62.7}
& \textbf{82.1}
& \textbf{49.8} \\

\quad SAPO
& 31.5
& \textbf{20.1}
& 35.0
& 75.4
& 39.4
& 59.7
& \textbf{82.1}
& 49.0 \\

\rowcolor{ModelGreen}
\quad \textbf{SAPO + DyCo-RL}
& \textbf{32.0}
& 19.7
& \textbf{35.3}
& \textbf{77.2}
& \textbf{44.5}
& \textbf{63.9}
& 81.1
& \textbf{50.5} \\

\quad GSPO
& \textbf{27.2}
& 19.7
& 34.0
& 76.6
& 35.8
& \textbf{62.4}
& 79.6
& 47.9 \\

\rowcolor{ModelGreen}
\quad \textbf{GSPO + DyCo-RL}
& 26.0
& \textbf{24.0}
& \textbf{34.1}
& \textbf{84.0}
& \textbf{37.4}
& 62.0
& \textbf{81.9}
& \textbf{49.9} \\

\midrule

\rowcolor{gray!15}
\multicolumn{9}{c}{\textbf{Qwen2.5-VL-7B}} \\

\quad \textit{Zero-Shot}
& 33.2
& 23.7
& 43.8
& 82.9
& 44.5
& 65.0
& 82.4
& 53.6 \\

\quad GRPO
& 38.7
& 26.6
& 47.4
& 78.0
& 46.1
& 69.2
& 83.3
& 55.6 \\

\rowcolor{ModelGreen}
\quad \textbf{GRPO + DyCo-RL}
& \textbf{40.5}
& \textbf{29.3}
& \textbf{49.5}
& \textbf{83.8}
& \textbf{48.5}
& \textbf{71.6}
& \textbf{84.9}
& \textbf{58.3} \\

\quad DAPO
& 36.4
& 23.0
& 44.5
& 79.2
& 42.3
& 70.0
& \textbf{86.3}
& 54.5 \\

\rowcolor{ModelGreen}
\quad \textbf{DAPO + DyCo-RL}
& \textbf{37.5}
& \textbf{24.3}
& \textbf{45.2}
& \textbf{85.8}
& \textbf{43.8}
& \textbf{71.0}
& 86.0
& \textbf{56.2} \\

\quad SAPO
& 38.5
& \textbf{26.6}
& 47.0
& 82.4
& 42.3
& \textbf{71.0}
& 85.6
& 56.2 \\

\rowcolor{ModelGreen}
\quad \textbf{SAPO + DyCo-RL}
& \textbf{40.7}
& 24.7
& \textbf{47.8}
& \textbf{84.3}
& \textbf{47.7}
& 69.2
& \textbf{86.0}
& \textbf{57.2} \\

\quad GSPO
& 39.4
& 26.6
& 42.4
& 75.8
& 46.5
& 71.4
& 81.6
& 54.8 \\

\rowcolor{ModelGreen}
\quad \textbf{GSPO + DyCo-RL}
& \textbf{39.9}
& \textbf{27.6}
& \textbf{46.5}
& \textbf{85.5}
& \textbf{48.3}
& \textbf{73.2}
& \textbf{84.8}
& \textbf{58.0} \\
\midrule

\rowcolor{gray!15}
\multicolumn{9}{c}{\textbf{Qwen3.5-27B}} \\









\quad \textit{Zero-Shot}
& 55.0 & 33.6 & 57.7 & 84.5 & 43.6 & 71.4 & 87.9
& 62.0 \\

\quad GRPO
& \textbf{61.4} & 39.8 & 57.9 & 82.6 & 45.2 & 73.1 & 88.5
& 64.1 \\

\rowcolor{ModelGreen}
\quad \textbf{GRPO + DyCo-RL}
& 61.2 & \textbf{41.1} & \textbf{61.0} & \textbf{86.4} & \textbf{49.0} & \textbf{74.3} & \textbf{89.1}
& \textbf{66.0} \\

\quad DAPO
& \textbf{60.5} & 38.5 & 59.0 & 83.8 & 45.4 & 71.0 & 87.1
& 63.6 \\

\rowcolor{ModelGreen}
\quad \textbf{DAPO + DyCo-RL}
& 59.3 & \textbf{40.1} & \textbf{59.9} & \textbf{85.6} & \textbf{48.8} & \textbf{72.9} & \textbf{89.4}
& \textbf{65.1} \\

\quad SAPO
& 58.7 & 36.2 & 58.4 & 85.0 & 45.4 & 71.3 & \textbf{88.4}
& 63.3 \\

\rowcolor{ModelGreen}
\quad \textbf{SAPO + DyCo-RL}
& \textbf{61.0} & \textbf{38.5} & \textbf{58.8} & \textbf{87.1} & \textbf{46.1} & \textbf{71.8} & \textbf{88.4}
& \textbf{64.5} \\

\quad GSPO
& 60.0 & 37.2 & 60.0 & 85.1 & \textbf{47.7} & 72.1 & 88.6
& 64.4 \\

\rowcolor{ModelGreen}
\quad \textbf{GSPO + DyCo-RL}
& \textbf{61.9} & \textbf{39.5} & \textbf{60.2} & \textbf{87.3} & 47.0 & \textbf{75.2} & \textbf{88.9}
& \textbf{65.7} \\

\bottomrule
\end{tabular}
\end{adjustbox}
\vspace{-15pt}
\end{table*}
Table~\ref{tab:main} presents evaluation results on seven benchmarks across three backbone configurations (Qwen2.5-VL-3B/7B and Qwen3.5-27B) and four RLVR algorithms. DyCo-RL improves the average performance of every evaluated baseline across all three configurations, supporting its applicability across model scales and optimization algorithms.

\vspace{-10pt}
\paragraph{Algorithm-Agnostic Generality.}
DyCo-RL yields average performance gains across GRPO, DAPO, SAPO, and GSPO without requiring algorithm-specific modifications. Despite differences in their optimization objectives, the improvements hold under every evaluated backbone configuration. This transferability across clip-based, sequence-level, and soft-adaptive optimization paradigms suggests that alignment-guided advantage reweighting is compatible with diverse RLVR objectives and can serve as a broadly applicable enhancement within the evaluated pipelines.

\vspace{-10pt}
\paragraph{Balanced Improvements Across Reasoning and Perception.}
The gains span both mathematical and visual-centric reasoning benchmarks. Improvements on visual tasks such as HallusionBench and MME are accompanied by gains on MathVision and MathVerse in multiple configurations. Although individual benchmarks exhibit occasional regressions, the overall results support the value of jointly considering visual and textual demands when optimizing multimodal reasoning.

\vspace{-10pt}
\paragraph{Scalability Across Model Sizes and Generations.}
Beyond Qwen2.5-VL-3B, DyCo-RL improves average performance on Qwen2.5-VL-7B and Qwen3.5-27B across all four algorithms. 
These improvements extend across model sizes and backbone generations, supporting the applicability of role-aware advantage reweighting beyond a single backbone configuration.

\vspace{-7pt}
\subsection{Comparison with Visual Reasoning Methods}
\vspace{-7pt}
Beyond general RLVR algorithms, we compare DyCo-RL with Attention Intervention and two representative methods with publicly available implementations, PAPO-G and VPPO, using Qwen2.5-VL-3B as the common backbone. Attention Intervention applies the head-wise enhancement in Eq.~(5) with $\lambda=0.2$ to the zero-shot model at inference time, without RL fine-tuning. Target modalities are determined by the Fisher--Rao roles in Eq.~(8). PAPO-G~\citep{wang2025perception} augments GRPO with a KL-based implicit perception loss. VPPO~\citep{huang2025spotlight} reweights trajectory-level advantages by visual dependency and restricts policy updates to perceptually pivotal tokens. These comparisons evaluate role-aware advantage reweighting against inference-time intervention, perception-focused objectives, and visually guided token-level optimization.

As shown in Table~\ref{tab:visual_reasoning_methods}, DyCo-RL achieves the highest average score (49.3), exceeding PAPO-G (47.9), VPPO (48.4), and Attention Intervention (48.1). It outperforms PAPO-G on all seven benchmarks, including gains of 3.1 points on WeMath and 1.9 points on MMBench. It surpasses VPPO on six benchmarks, with gains of 1.1 points on MMBench, 2.2 points on MME, and 1.6 points on LogicVista. These results support incorporating dynamic visual--textual coordination into training, with benefits over the evaluated inference-time and perception-focused methods.

\begin{table}[t]
\vspace{-10pt}
\scriptsize
\centering
\caption{Comparison with  visual reasoning methods on Qwen2.5-VL-3B.}
\label{tab:visual_reasoning_methods}
\setlength{\tabcolsep}{4.5pt}
\renewcommand{\arraystretch}{1.12}
\begin{adjustbox}{max width=\linewidth}
\begin{tabular}{lcccccccc}
\toprule
\textbf{Method (Qwen2.5-VL-3B)}
& \textbf{WeMath}
& \textbf{MathVision}
& \textbf{MathVerse}
& \textbf{MMBench}
& \textbf{LogicVista}
& \textbf{HallusionBench}
& \textbf{MME}
& \textbf{Avg.} \\
\midrule
Zero-Shot
& 20.0 & 23.0 & 33.8 & 71.4 & 34.5 & 61.3 & 76.9 & 45.8 \\

GRPO
& 23.0 & 21.4 & 34.4 & 72.8 & 39.8 & 60.8 & 79.9 & 47.4 \\

Attention Intervention
& 24.4 & 21.7 & 34.5 & 74.0 & 40.3 & \textbf{62.7} & 79.0 & 48.1 \\

PAPO-G
& 22.8 & 21.4 & 34.6 & 74.3 & 40.3 & 61.8 & 80.3 & 47.9 \\

VPPO
& \textbf{26.9} & 22.0 & 35.8 & 75.1 & 39.6 & 61.2 & 78.5 & 48.4 \\

\rowcolor{ModelGreen}
\textbf{GRPO + DyCo-RL}
& 25.9 & \textbf{22.4} & \textbf{36.0} & \textbf{76.2}
& \textbf{41.2} & 62.4 & \textbf{80.7} & \textbf{49.3} \\
\bottomrule
\end{tabular}
\end{adjustbox}
\end{table}

\begin{table*}[t]
\vspace{-10pt}
\scriptsize
\centering
\caption{Effectiveness of Fisher--Rao Role Assignment compared to alternative assignment strategies.}

\label{tab:ablation_role}

\setlength{\tabcolsep}{5pt}
\renewcommand{\arraystretch}{1.15}

\begin{adjustbox}{max width=\linewidth}
\begin{tabular}{lcccccccc}
\toprule

\textbf{Method}
& \textbf{WeMath}
& \textbf{MathVision}
& \textbf{MathVerse}
& \textbf{MMBench}
& \textbf{LogicVista}
& \textbf{HallusionBench}
& \textbf{MME}
& \textbf{Avg.} \\

\midrule

GRPO (Baseline)
& 23.0
& 21.4
& 34.4
& 72.8
& 39.8
& 60.8
& 79.9
& 47.4 \\

\quad + Random
& 23.5
& 20.1
& 33.1
& 72.4
& 36.7
& 59.7
& 79.3
& 46.4 \\

\quad + Reverse
& 24.2
& 19.4
& 34.0
& 70.6
& 36.0
& 58.6
& 78.0
& 45.8 \\

\quad + Entropy
& 25.0
& 21.1
& 34.4
& 74.5
& 40.3
& 61.2
& \textbf{81.1}
& 48.2 \\

\quad + KL Divergence
& 24.7
& 21.4
& 35.4
& 75.3
& 40.7
& \textbf{62.5}
& 80.4
& 48.6 \\

\rowcolor{ModelGreen}
\quad \textbf{+ DyCo-RL (Ours)}
& \textbf{25.9}
& \textbf{22.4}
& \textbf{36.0}
& \textbf{76.2}
& \textbf{41.2}
& 62.4
& 80.7
& \textbf{49.3} \\

\bottomrule
\end{tabular}
\end{adjustbox}
\vspace{-10pt}
\end{table*}

\begin{table*}[t]
\vspace{-10pt}
\scriptsize
\centering

\caption{Effectiveness of Alignment-Guided Reweighting.}

\label{tab:ablation_1}

\setlength{\tabcolsep}{5pt}
\renewcommand{\arraystretch}{1.15}

\begin{adjustbox}{max width=\linewidth}
\begin{tabular}{lcccccccc}
\toprule

\textbf{Method}
& \textbf{WeMath}
& \textbf{MathVision}
& \textbf{MathVerse}
& \textbf{MMBench}
& \textbf{LogicVista}
& \textbf{HallusionBench}
& \textbf{MME}
& \textbf{Avg.} \\

\midrule
GRPO (Baseline)
& 23.0
& 21.4
& 34.4
& 72.8
& 39.8
& 60.8
& 79.9
& 47.4 \\

\quad + Visual Attn Score ($\tilde{r}_t^{\text{vis}}$)
& 24.5
& \textbf{23.0}
& 32.1
& 74.4
& 37.4
& 62.0
& 78.1
& 47.4 \\

\quad + Text Attn Score ($\tilde{r}_t^{\text{txt}}$)
& \textbf{26.0}
& 22.4
& 35.4
& 70.7
& 37.8
& 59.4
& 78.4
& 47.2 \\

\rowcolor{ModelGreen}
\quad \textbf{+ DyCo-RL (Ours)}
& 25.9
& 22.4
& \textbf{36.0}
& \textbf{76.2}
& \textbf{41.2}
& \textbf{62.4}
& \textbf{80.7}
& \textbf{49.3} \\
\bottomrule
\end{tabular}
\end{adjustbox}
\vspace{-10pt}
\vspace{-5pt}
\end{table*}
\vspace{-7pt}
\subsection{Ablation Studies}
\vspace{-5pt}

To systematically isolate the performance gains introduced by DyCo-RL, we independently ablate its two core modules: the token role assignment strategy and the alignment-guided reweighting mechanism. Ablation experiments are conducted on the Qwen2.5-VL-3B model using the GRPO baseline. Additional ablation results are provided in Appendix~\ref{app:abl}.
\vspace{-10pt}
\paragraph{Effectiveness of Fisher--Rao Role Assignment.}
We compare DyCo-RL against four alternative token role assignment strategies (Table~\ref{tab:ablation_role}).
\textit{Random} assigns roles arbitrarily and lowers the average score by 1.0 points relative to GRPO, while \textit{Reverse} inverts our assignments and yields the lowest average score. These controls support the importance of informed role assignment and indicate that reversing the assignment direction reduces the effectiveness of reweighting.

We further compare two heuristic baselines: \textit{Entropy}, which measures the uncertainty of the within-modality attention distribution at a single step, and \textit{KL}, which captures changes between consecutive distributions. Entropy provides a static snapshot, whereas KL incorporates temporal changes but is asymmetric. Fisher--Rao instead provides a symmetric, geometrically principled distance on the probability simplex, characterizing how attention is redistributed within each modality between consecutive decoding steps.
DyCo-RL achieves the highest average performance among the evaluated variants, supporting the use of Fisher--Rao-based role assignment for token-level advantage reweighting.
\vspace{-10pt}
\paragraph{Effectiveness of Alignment-Guided Reweighting.}
To validate the conditional formulation of our alignment score $s_t$ in Eq.~\eqref{eq_attention}, we replace it with two modality-biased alternatives while keeping all other components identical. Specifically, rather than dynamically switching the scoring function based on a token's assigned role ($\mathcal{D}^{\text{vis}}$ or $\mathcal{D}^{\text{txt}}$),
we use a fixed modality's min--max normalized attention mass for all tokens in $\mathcal{D}$: \textit{(i) + visual attention score} (setting $s_t = \tilde{r}_t^{\text{vis}}$), and \textit{(ii) + text attention score} (setting $s_t = \tilde{r}_t^{\text{txt}}$). As shown in Table~\ref{tab:ablation_1}, optimizing either modality alone can improve certain benchmarks but degrades others, and neither consistently surpasses GRPO. This indicates that one-sided modality optimization is insufficient. In contrast, DyCo-RL jointly accounts for both modalities and selectively reweights tokens according to their functional roles, yielding the highest overall performance.

\vspace{-10pt}
\section{Discussion and Analysis}
\label{sec:analysis}
\vspace{-9pt}
Beyond empirical gains, understanding \textit{why} DyCo-RL succeeds requires a deeper look into the model's internal representations. To this end,
we sample 200 generated instances from both the vanilla GRPO baseline and GRPO+DyCo-RL in MathVerse and MathVision. Because the two models generate distinct reasoning trajectories, we independently annotate their tokens into visually-oriented and text-oriented roles following the diagnostic protocol detailed in Section~\ref{4.1}, with annotators blinded to model identity. Based on this curated evaluation set, we investigate how our alignment-guided reweighting fundamentally reshapes cross-modal reasoning dynamics.

\begin{figure*}[h]
\vspace{-10pt}
    \centering
    \includegraphics[width=1\linewidth]{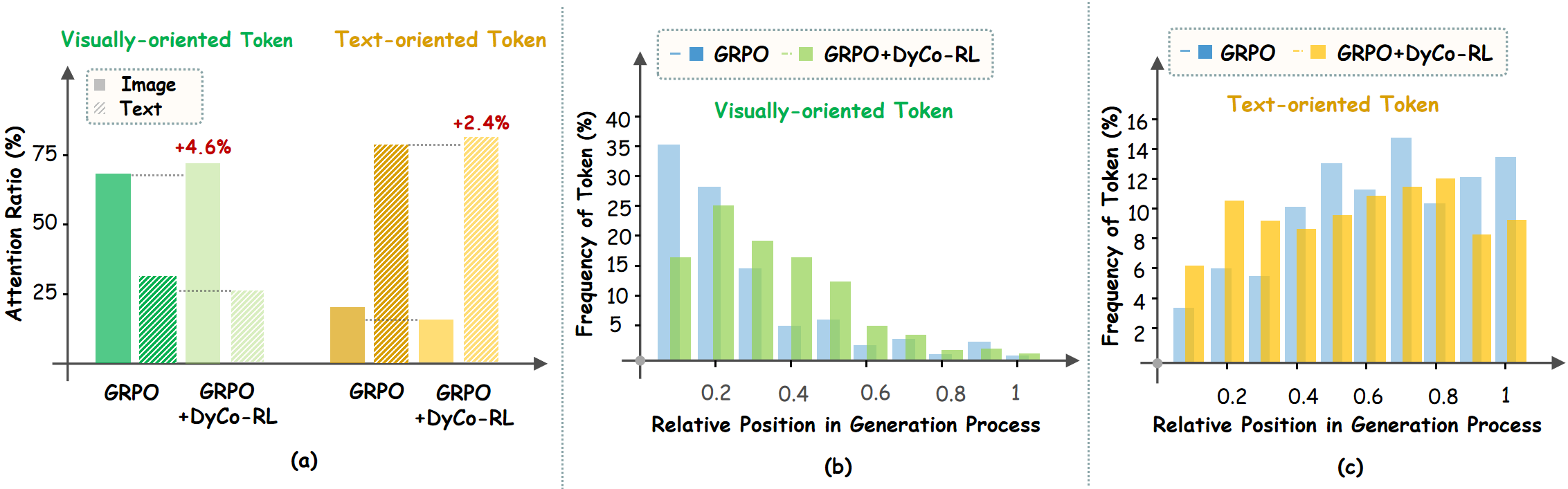}
    \vspace{-20pt}
 \caption{Effectiveness of DyCo-RL in dynamic cross-modal coordination. (a) Alignment between assigned token roles and modality attention ratios; (b) Distribution of visually-oriented tokens across the generation process; (c) Distribution of text-oriented tokens across the generation process.}

    \label{fig:analysis}
    \vspace{-10pt}
\end{figure*}

\vspace{-5pt}
\paragraph{Strengthening Role-Aligned Attention.}
We first evaluate the attention allocations across the annotated tokens. Attention ratios are computed as in Fig.~\ref{fig:relevant}. As illustrated in Fig.~\ref{fig:analysis}(a), DyCo-RL strengthens the alignment between a token's functional role and its actual modality attention. Specifically, compared with GRPO, DyCo-RL exhibits a higher image attention ratio for visually-oriented tokens and a higher text attention ratio for text-oriented tokens. The shift is partly expected by construction, since the objective rewards role-aligned attention. This trend confirms that concentrating the learning signal on role-aligned tokens shifts attention toward the designated modality.


\vspace{-10pt}

\paragraph{Reshaping Temporal Dynamics.}
Fig.~\ref{fig:analysis}(b--c) shows the temporal distribution of each token role, measured as the fraction of its tokens in each relative-position bin.
Under GRPO, visually-oriented tokens are concentrated at the beginning, with a small proportion appearing after the midpoint, while text-oriented tokens become increasingly frequent toward the end.
DyCo-RL reduces this separation: visually-oriented tokens have a greater presence in the middle stages, while text-oriented tokens appear earlier and are distributed more evenly.
This more interleaved distribution suggests a reduced tendency to ``perceive first, reason later'' and, together with the stronger role--attention alignment in Fig.~\ref{fig:analysis}(a), supports dynamic cross-modal coordination as the CoT unfolds.

\vspace{-8pt}
\section{Conclusion}
\label{sec:conclusion}
\vspace{-8pt}
We identify cross-modal coordination breakdowns as a contributing factor to visual reasoning errors in RLVR, where visually- and text-oriented tokens can under-attend to their designated modalities. To address this, we propose DyCo-RL, an algorithm-agnostic plugin that assigns functional token roles via Fisher--Rao attention distance and dynamically reweights advantages based on role-attention alignment. Extensive evaluations show that DyCo-RL improves the aggregate performance of four representative baselines across seven diverse benchmarks,
supporting explicit token-level coordination as an effective approach to improving multimodal reasoning.

\section*{AI Usage Statement}

An LLM assisted with manuscript and code revision, and literature checking. The authors retain full responsibility for the research decisions, experimental results, and final manuscript.


\bibliographystyle{plainnat}
\bibliography{custom}
\newpage
\newpage
\appendix
\section{Overview of Appendix}

\begin{itemize}
    \item \ref{app:limitation}: \textbf{Limitations}.
    \item \ref{app:impact}: \textbf{Broader Impact}.
    \item \ref{app:alg}:
     \textbf{Details of Experimental Setup.}

    \item \ref{appendix:settings}: \textbf{Implementation Details}.

    \item \ref{app:hal_analysis}: \textbf{Diagnostic Annotation Protocol}.
    \item  \ref{app:position_analysis}: \textbf{Additional Correlation Analysis}.
     \item \ref{app:intervention_controls}: \textbf{Details of Causal Intervention Protocol and Additional Controls}.

    \item \ref{app:role_analysis}: \textbf{Additional Analysis of Token Role Assignment}.
    \item \ref{app:abl}: \textbf{Additional Ablation Studies}.
    \item \ref{app:multiseed}: \textbf{Multi-Seed Evaluation}.
    \item \ref{app:overhead}: \textbf{Analysis of Computational Overhead}.
    \item \ref{app:dyn}: \textbf{Analysis of Training Dynamics}.

    \item \ref{app:gen}: \textbf{Extended Generalization Analysis}.
    
    \item \ref{app:prompt}: \textbf{Prompt Template}.
    \item \ref{app:vis}: \textbf{More Visualization Results}.
    
\end{itemize}

\section{Limitations}
\label{app:limitation}
Our work has several limitations. First, although DyCo-RL introduces no additional trainable parameters, extracting per-token attention statistics and computing Fisher--Rao distances during rollout incur additional training time and GPU memory overhead. To limit memory usage, we extract attention statistics only from the last decoder layer, leaving the benefits of incorporating signals from other layers unexplored. Second, due to limited computational resources, we train for a single epoch on ThinkLite-hard-11K and evaluate only Qwen-family backbones. Generalization to other training datasets and model families remains to be verified.
\section{Broader Impact}
\label{app:impact}

The core mechanism of DyCo-RL, which leverages within-modality attention restructuring to assign functional roles and perform alignment-guided advantage reweighting, extends naturally beyond visual mathematical reasoning. Cross-modal coordination failures, where a model struggles to dynamically alternate between visual grounding and textual deduction, are a pervasive bottleneck across long-chain multimodal tasks. Whenever an objective requires interleaving perceptual evidence with contextual or symbolic inference, our role-aware token-level optimization becomes directly applicable.

Representative scenarios include visual document and chart understanding~\citep{he2025distill}, long-video reasoning~\citep{zhang2025thinking}, remote sensing reasoning~\citep{shu2026terrascope}, and medical image analysis~\citep{shu2025flemingvluniversalmedicalvisual}. All these domains require continuous dynamic coordination between perceptual grounding and higher-level reasoning. In such settings, coordination breakdowns frequently manifest as hallucinated details, temporally inconsistent claims, or derivation steps that drift away from the underlying visual evidence.

More broadly, standard RLVR paradigms treat cross-modal coordination as a mere byproduct of sequence-level reward broadcasting. By elevating it to an explicit token-level optimization signal, 
DyCo-RL may benefit multimodal reasoning applications through improved cross-modal coordination. However, role-aligned attention does not guarantee faithful or correct reasoning, and models may still produce misleading outputs. Application-specific validation remains necessary before deployment. Since the Fisher--Rao assignment derives entirely from intrinsic attention geometry and the reweighting strategy remains algorithm-agnostic, this principle integrates seamlessly with existing reward designs and modern policy optimization baselines such as GRPO. 

We envision role-aware token-level optimization as a core component for trustworthy multimodal AI systems deployed in high-stakes domains (e.g., healthcare, autonomous systems, and scientific discovery). In these critical fields, the step-by-step fidelity of the reasoning process is just as paramount as the final-answer correctness.

\section{Details of Experimental Setup}
\label{app:alg}

To provide a comprehensive overview of our evaluation framework, we detail the representative baseline algorithms and the selected benchmarks below.

\paragraph{Baseline Algorithms.}
We evaluate DyCo-RL against four reinforcement learning baselines that span the recent evolution of policy optimization for reasoning-oriented large language models. \textbf{GRPO}~\citep{shao2024deepseekmathpushinglimitsmathematical} relies on intra-group reward normalization to eliminate the auxiliary critic model. \textbf{DAPO}~\citep{yu2025dapoopensourcellmreinforcement} extends this by introducing asymmetric clipping and dynamic rollout strategies to stabilize long-chain reasoning. \textbf{SAPO}~\citep{gao2025softadaptivepolicyoptimization} utilizes a smooth, sigmoid-based soft gating mechanism with asymmetric temperatures to address optimization discontinuities. Finally, \textbf{GSPO}~\citep{zheng2025groupsequencepolicyoptimization}
uses a length-normalized sequence-level importance ratio
and sequence-level clipping. To incorporate DyCo-RL's
token-specific advantages, we adopt its GSPO-token variant,
whose stop-gradient construction preserves the sequence-level
ratio while routing gradients through the corresponding
token likelihoods. The reweighted advantages are applied
to the token-wise surrogate terms before clipping and
subsequent averaging.

\paragraph{Evaluation Benchmarks.}
We evaluate DyCo-RL on seven benchmarks covering mathematical reasoning and visual-centric reasoning to assess visual grounding and reasoning consistency. We follow the benchmark-specific evaluation procedures implemented in VLMEvalKit~\citep{duan2024vlmevalkit}, using the splits and metrics specified below.

\noindent\textbf{Mathematical Reasoning.}
\begin{itemize}
  \item \textbf{WeMath}~\citep{qiao2025we}: A hierarchical visual mathematics benchmark that decomposes composite problems into subproblems associated with individual knowledge concepts, enabling diagnosis of knowledge gaps and reasoning failures. We evaluate on its \texttt{testmini} split, containing $1{,}740$ questions, and report the Strict score computed by VLMEvalKit.
  \item \textbf{MathVision}~\citep{wang2024measuring}: A visual mathematical reasoning benchmark comprising $3{,}040$ competition problems across 16 mathematical disciplines. Our evaluation uses the 304-question \texttt{testmini} subset, registered as \texttt{MathVision\_MINI} in VLMEvalKit, and reports final-answer accuracy.
  \item \textbf{MathVerse}~\citep{zhang2024mathverse}: A diagram-based mathematical reasoning benchmark with controlled variations in the information provided through text and images. We evaluate on \texttt{MathVerse\_MINI}, which contains 788 problems presented in five versions: Text Dominant, Text Lite, Vision Intensive, Vision Dominant, and Vision Only, yielding $3{,}940$ evaluation instances. We compute final-answer accuracy for each version.
\end{itemize}

\noindent\textbf{Visual-centric Reasoning.}
\begin{itemize}
  \item \textbf{MMBench}~\citep{liu2024mmbench}: A multiple-choice multimodal benchmark covering diverse perception and reasoning abilities. We use the English development split of v1.0 (\texttt{MMBench\_DEV\_EN}) and report circular accuracy, which evaluates consistency across permutations of the answer options.
  \item \textbf{LogicVista}~\citep{xiao2024logicvista}: A visual logical reasoning benchmark covering inductive, deductive, numerical, spatial, and mechanical reasoning. We evaluate the 447 questions provided by VLMEvalKit under \texttt{LogicVista} and report overall answer accuracy.
  \item \textbf{HallusionBench}~\citep{Guan_2024_CVPR}: A diagnostic benchmark designed to probe visual illusions and language-prior-induced hallucinations through paired image--question controls. We evaluate the 951 instances provided by VLMEvalKit and report answer-level accuracy (aAcc), the proportion of individual responses that correctly answer the corresponding yes/no questions.
  \item \textbf{MME}~\citep{fu2023mme}: A comprehensive evaluation suite covering 14 perception and cognition tasks, including object recognition, OCR, and commonsense reasoning. We use the standard \texttt{MME} evaluation set and its official scoring procedure: each task receives the sum of answer accuracy (Acc) and paired-question accuracy (Acc+), with task scores aggregated within the perception and cognition categories.
\end{itemize}

\noindent\textbf{Evaluation Protocol.}
We perform evaluation using the VLM-R1 interface in VLMEvalKit~\citep{duan2024vlmevalkit}. Prompts request reasoning and final answers within \texttt{<think>} and \texttt{<answer>} tags, respectively. The answer-tag content is extracted when available; otherwise, the full response is retained. Each baseline and its DyCo-RL counterpart share the same prompting, answer-processing, and decoding settings, with a decoding temperature of 0.1.

For WeMath, MathVision, MathVerse, and LogicVista, the configured answer-extraction model is \texttt{gpt-4o-mini-2024-07-18}, which also performs answer judging where required by the benchmark-specific evaluator. For MMBench, HallusionBench, and MME, VLMEvalKit uses rule-based answer extraction with \texttt{gpt-3.5-turbo-0125} configured as the auxiliary extractor when rule-based extraction is insufficient. For MME, we normalize the combined perception and cognition score $S_{\mathrm{MME}}$ to a 0--100 scale as $100 \times S_{\mathrm{MME}}/2800$. The overall average is the arithmetic mean of the seven benchmark scores.

We additionally compare answer-extraction failure rates in
Table~\ref{tab:extraction_failure}, counting answers that remain
unresolved after benchmark-specific answer processing.
For each model scale, we average these rates equally across
the four RLVR algorithms and a common set of benchmarks,
comparing each algorithm with and without DyCo-RL. The comparable average extraction failure rates suggest that
the observed performance gains are unlikely to be primarily
driven by differences in answer extractability.
\vspace{-10pt}
\begin{table}[H]
\scriptsize
\centering
\caption{Average answer-extraction failure rates (\%) across
four RLVR algorithms and seven benchmarks.}
\label{tab:extraction_failure}
\setlength{\tabcolsep}{10pt}
\renewcommand{\arraystretch}{1.15}
\begin{tabular}{lcc}
\toprule
\textbf{Model}

& \textbf{Algorithms (Avg.)}
& \textbf{Algorithms + DyCo-RL (Avg.)} \\
\midrule
Qwen2.5-VL-3B & 2.02 & 2.11 \\
Qwen2.5-VL-7B & 1.67 & 1.31 \\
\bottomrule
\end{tabular}
\end{table}

\section{Implementation Details}
\label{appendix:settings}
All experiments are conducted within the VLM-R1 training framework~\citep{shen2025vlm}, which serves as the base training paradigm for all models in this work. And as summarized in Table~\ref{tab:hyperparams}, all models are trained for a single epoch on the ThinkLite-hard-11K dataset, which comprises 11,031 complex reasoning instances. During rollout generation, we sample 4 responses per prompt using a temperature of 0.9. The training objective is guided by a composite reward signal, integrating a binary accuracy reward to verify mathematical correctness and a format reward to enforce structural compliance. 

Unlike standard parameter-efficient fine-tuning, we leave the vision tower entirely \emph{unfrozen} to enable full-parameter cross-modal optimization. Across all evaluated baseline algorithms, we employ the AdamW optimizer with a peak learning rate of $1\times10^{-6}$ and a linear decay schedule. To stabilize the training dynamics, we apply gradient checkpointing and cap the global gradient norm at 1.0. For the specific hyperparameter configurations of each RL method (e.g., clipping bounds or KL coefficients),
we follow the settings of the corresponding official implementations unless explicitly stated otherwise.

All experiments are executed in \texttt{bfloat16} precision, heavily relying on DeepSpeed ZeRO-3 optimization (without CPU offloading) to manage memory efficiency. We configure the per-device batch size to 4 and accumulate gradients over 2 steps, yielding a global batch size of 64. The entire training pipeline is deployed on a single server equipped with 8 $\times$ NVIDIA A100 (80GB) GPUs. Rollouts are generated using Hugging Face Transformers' \texttt{generate}.
Baseline training for one epoch requires approximately
54 GPU-hours for Qwen2.5-VL-3B, 72 GPU-hours for Qwen2.5-VL-7B,
and 200 GPU-hours for Qwen3.5-27B, averaged across the evaluated
baseline algorithms. Additional analysis of computational overhead is reported in
Appendix~\ref{app:overhead}.

\begin{table}[t]
\scriptsize
  \centering
  \caption{Key hyperparameters.}
  \label{tab:hyperparams}
\setlength{\tabcolsep}{8pt}
\renewcommand{\arraystretch}{1.15}
  \begin{adjustbox}{max width=\linewidth}
\begin{tabular}{lc}
  \toprule
  \textbf{Hyperparameter} & \textbf{Value} \\
  \midrule
  Optimizer                       & AdamW \\
  Learning Rate                   & $1 \times 10^{-6}$ \\
  LR Schedule                     & linear \\
  Epochs                          & 1 \\
  Freeze Vision Tower             & No \\
  Global Batch Size (rollout sequences) & 64 \\
  Per-Device Batch Size           & 4 \\
  Gradient Accumulation           & 2 \\
  GPUs per Node                   & 8 \\
  Rollouts per Prompt ($G$)       & 4 \\
  Rollout Temperature             & 0.9 \\
  Rollout Top-$p$                 & 1.0 \\
  Max Response Length              & 2048 \\
  Reward Signal                   & accuracy + format \\
  Precision                       & bfloat16 \\
  Gradient Clipping               & 1.0 \\
  DeepSpeed Stage                 & ZeRO-3 \\
  \bottomrule
  \end{tabular}
\end{adjustbox}
\end{table}

\section{Diagnostic Annotation Protocol}
\label{app:hal_analysis}

To ground our analysis of visual reasoning breakdowns in concrete model behavior, we construct a token-level annotated error set using the Qwen2.5-VL-3B model optimized with our standard GRPO baseline. We uniformly sample 200 incorrect rollouts from MathVerse and MathVision, two benchmarks covering diverse forms of visual mathematical reasoning, including diagram understanding, geometric reasoning, symbolic derivation, and chart interpretation. Annotators are asked to identify, for each generated token span, both the source of failure (visual grounding versus textual reasoning) and whether the corresponding operation is executed correctly. This dual-axis evaluation yields a comprehensive four-way taxonomy.

\paragraph{Annotation Principle.}
Annotations are assigned according to a token's \textit{functional role} within the local reasoning trajectory rather than its isolated lexical meaning. Specifically, annotators determine whether a token primarily contributes to image-grounded perception or text-based reasoning under the current decoding context. To improve annotation consistency, labels are first assigned at the semantic-span level and subsequently projected back to individual tokens. Connective tokens are categorized according to the functional role of their subsequent semantic span. For example, the prefix ``since angle ADE is $80^\circ$'' is labeled as \textit{visually-oriented} because the span performs image-grounded perception, whereas ``since Eq.~(3) implies'' is categorized as \textit{text-oriented} reasoning. 

\paragraph{Token Taxonomy.}
Only tokens within spans expressing visual grounding or
reasoning operations are annotated, and each annotated token is assigned to exactly one of the following four categories:

\begin{itemize}
    \item \textbf{Correct Visually-oriented Token ($\mathcal{P}^{r}$).} Tokens that accurately extract or describe image-grounded information. Examples include correctly reading numerical values from a chart, identifying existing geometric relations, or accurately recognizing objects from the visual input.
    \item \textbf{Incorrect Visually-oriented Token ($\mathcal{P}^{e}$).} Tokens that attempt to describe visual content but directly contradict the provided image. This includes misread numerical values, hallucinated visual elements, or erroneous spatial descriptions.
    \item \textbf{Correct Text-oriented Token ($\mathcal{R}^{r}$).} Tokens that perform logically valid inference operations conditioned strictly on the preceding context. This encompasses correct algebraic manipulations, valid theorem applications, or intermediate symbolic deductions without requiring new visual evidence.
    \item \textbf{Incorrect Text-oriented Token ($\mathcal{R}^{e}$).} Tokens corresponding to logically inconsistent or unsupported reasoning steps. Examples include self-contradictory mathematical conclusions, invalid symbolic derivations, or abrupt claims lacking traceable antecedents in the reasoning history.
\end{itemize}

\paragraph{Annotation Procedure and Agreement.}
Each sample is independently evaluated by two external annotators possessing graduate-level mathematical training. While annotators are provided with the input image, question, full model response, and reference answer, they are explicitly instructed to label tokens based on the internal consistency and visual faithfulness of the generated trajectory, rather than retrofitting labels based on final-answer correctness. Any labeling discrepancies between the two primary annotators are resolved by a third senior referee. These manual labels are used only for diagnostic analysis and are not involved in DyCo-RL training.

To quantitatively evaluate annotation reliability, we compute Cohen's $\kappa$ at the token-span level before the final adjudication. The resulting agreement score ($\kappa = 0.85$)
indicates high inter-annotator agreement. The majority of the initial disagreements (approximately 3\% of all annotated tokens) are concentrated around ambiguous transitions between perception and reasoning spans.

\paragraph{Annotation Statistics.}
Table~\ref{tab:annotation_statistics} reports the token counts for the four annotation groups. The correct-token groups, $\mathcal{P}^r$ and $\mathcal{R}^r$, contain tokens from locally correct spans within the same 200 erroneous rollouts.

\begin{table}[htbp]
\scriptsize
    \centering
    \vspace{10pt}
    \caption{Token counts in the 200 annotated erroneous rollouts.}
    \label{tab:annotation_statistics}
    \setlength{\tabcolsep}{8pt}
\renewcommand{\arraystretch}{1.15}
    \begin{tabular}{lllr}
        \toprule
        Group & Functional role & Correctness & Tokens \\
        \midrule
        $\mathcal{P}^r$ & Visually-oriented & Correct   & 1,533 \\
        $\mathcal{P}^e$ & Visually-oriented & Erroneous &   457 \\
        $\mathcal{R}^r$ & Text-oriented     & Correct   & 3,624 \\
        $\mathcal{R}^e$ & Text-oriented     & Erroneous & 1,261 \\
        \midrule
        \multicolumn{3}{l}{Total} & 6,875 \\
        \bottomrule
    \end{tabular}
\end{table}

\section{Additional Correlation Analysis}
\label{app:position_analysis}

To examine whether the attention gaps in Fig.~\ref{fig:relevant} can be explained by differences in generation position, we compare correct and erroneous tokens at matched generation positions, separately for visually-oriented and text-oriented roles. We use the modality-specific attention ratio $r_t^m/(r_t^{\mathrm{vis}}+r_t^{\mathrm{txt}})$, with $m=\mathrm{vis}$ for visually-oriented tokens and $m=\mathrm{txt}$ for text-oriented tokens. For each role, we report the difference between correct and erroneous tokens. Positive values indicate greater attention to the role-relevant modality for correct tokens.

\begin{table}[htbp]
\scriptsize
    \centering
    \caption{Attention ratio differences between correct and erroneous tokens at matched generation positions. }
    \label{tab:position_control}
    \setlength{\tabcolsep}{8pt}
\renewcommand{\arraystretch}{1.15}
    \begin{tabular}{llr}
        \toprule
        Functional role & Target modality & Difference \\
        \midrule
        Visually-oriented ($\mathcal{P}$) & Image           & $+0.13$ \\
        Text-oriented ($\mathcal{R}$)     & Textual context & $+0.12$ \\
        \bottomrule
    \end{tabular}
\end{table}

As shown in Table~\ref{tab:position_control}, the differences remain positive for both roles after position matching: correct visually-oriented tokens allocate more attention to the image, while correct text-oriented tokens allocate more attention to the textual context. These descriptive results preserve the direction of the attention--correctness association observed in Fig.~\ref{fig:relevant}.

\section{Details of Causal Intervention Protocol and Additional Controls}
\label{app:intervention_controls}

We conduct diagnostic attention interventions on the erroneous responses annotated in Appendix~\ref{app:hal_analysis}. These experiments examine whether directing attention toward the modality associated with an erroneous span can recover the final answer.

\paragraph{Span-Level Replay and Stopping Rule.}
Each annotated erroneous span is evaluated in an independent replay, conditioned on the original image, question, and response prefix preceding that span. The prefix is fixed to the original response, while tokens from the beginning of the target span onward are generated by the model. Intervention starts with the decoding step that predicts the first token of the span.

After each intervened step, we compare the generated token with the original token at the corresponding position. If they match and the span has not ended, intervention continues at the next step. At the first mismatch, the newly generated token is retained and intervention is disabled from the next step onward. If all generated tokens match the original span, intervention ends after its last token. In either case, generation continues without further attention intervention until the response is complete. Labels from the original response are not transferred to the divergent continuation, and no later annotated spans are intervened on within the same replay. This stopping rule confines the use of the original annotation to decoding steps whose preceding token sequence still matches the original response. All replays use greedy decoding with identical settings across
conditions. The no-intervention control ($\lambda=0$) is
evaluated using the same replay protocol and yields a recovery
rate of 0\%.

\paragraph{Span Statistics and Recovery Rate.}
The number of erroneous spans per response is small in our diagnostic set, with all 200 responses containing at least one visually-oriented and one text-oriented erroneous span. For the visually-oriented intervention set, 186 of the 200 responses contain one erroneous span and the remaining 14 contain two. For the text-oriented intervention set, 174 responses contain one erroneous span and the remaining 26 contain two. Thus, 93\% and 87\% of the responses in the visual and textual intervention sets, respectively, contain only a single erroneous span, while the remaining responses contain at most two. 

Recovery rate is computed at the sample level. For each intervention condition and strength, a sample is counted as recovered if at least one independent span-level replay yields a correct final answer, and each sample is counted only once. The same aggregation rule is used for all control conditions. Importantly, replay attempts are matched across intervention conditions for each response; responses containing two annotated spans therefore receive the same number of replay attempts under Ours and the corresponding control conditions. Consequently, differences between intervention strategies are not driven by unequal replay budgets across conditions.

\paragraph{Causal Intervention Controls.}
We compare the role-aware intervention (Ours) with three controls
that vary the intervention location or target modality using the 200 erroneous rollouts we annotated.
No-label retains the intervention locations used by Ours but
randomly selects the modality to enhance, without using the
annotated functional roles.
Random instead randomizes the intervention locations while
retaining the designated target modality: visual attention in
the visually-oriented intervention experiment and textual
attention in the text-oriented intervention experiment.
Wrong-Role retains the annotated intervention locations but
enhances the modality opposite to the annotated role.
We evaluate $\lambda \in \{0.1,0.2,\ldots,0.8\}$.
For each condition and $\lambda$, recovery rate is the fraction
of evaluated samples for which at least one independent replay
yields a correct final answer, counting each sample only once.

\begin{table}[H]
\scriptsize
\centering
\caption{Controls for intervention on visually-oriented erroneous tokens.}
\vspace{2pt}
\label{tab:intervention_p}
\setlength{\tabcolsep}{8pt}
\renewcommand{\arraystretch}{1.15}
\begin{tabular}{ccccc}
\toprule
$P_\lambda$ & \textbf{Ours} & \textbf{No-label} & \textbf{Random} & \textbf{Wrong-Role} \\
\midrule
0.0 & 0.0\% & 0.0\% & 0.0\% & 0.0\% \\
0.1 & 7.0\% & 4.0\% & 5.5\% & 4.0\% \\
0.2 & 15.5\% & 16.0\% & 11.5\% & 7.5\% \\
0.3 & 13.0\% & 14.5\% & 10.5\% & 9.5\% \\
0.4 & 21.5\% & 13.0\% & 15.0\% & 8.0\% \\
0.5 & 32.5\% & 11.5\% & 20.0\% & 6.5\% \\
0.6 & 23.5\% & 7.0\% & 14.5\% & 6.5\% \\
0.7 & 28.0\% & 10.5\% & 14.5\% & 0.5\% \\
0.8 & 11.0\% & 2.0\% & 8.5\% & 0.5\% \\
\bottomrule
\end{tabular}
\end{table}

\begin{table}[H]
\scriptsize
\centering
\caption{Controls for intervention on text-oriented erroneous tokens.}
\vspace{2pt}
\label{tab:intervention_r}
\setlength{\tabcolsep}{8pt}
\renewcommand{\arraystretch}{1.15}
\begin{tabular}{ccccc}
\toprule
$R_\lambda$ & \textbf{Ours} & \textbf{No-label} & \textbf{Random} & \textbf{Wrong-Role} \\
\midrule
0.0 & 0.0\% & 0.0\% & 0.0\% & 0.0\% \\
0.1 & 14.5\% & 11.0\% & 9.0\% & 7.5\% \\
0.2 & 19.0\% & 15.0\% & 12.5\% & 4.0\% \\
0.3 & 26.0\% & 21.5\% & 19.5\% & 8.5\% \\
0.4 & 23.0\% & 13.5\% & 18.0\% & 9.0\% \\
0.5 & 21.5\% & 17.0\% & 19.0\% & 7.5\% \\
0.6 & 26.5\% & 14.0\% & 18.5\% & 4.0\% \\
0.7 & 24.5\% & 7.5\% & 17.0\% & 2.5\% \\
0.8 & 16.0\% & 1.5\% & 14.0\% & 1.0\% \\
\bottomrule
\end{tabular}
\end{table}

Tables~\ref{tab:intervention_p} and~\ref{tab:intervention_r}
show that the role-aware intervention achieves higher recovery
rates than the controls across most tested strengths.
Wrong-Role generally yields the lowest recovery rates.
Random usually performs below Ours, suggesting that the
intervention location matters even when the target modality
is retained. No-label remains competitive at small strengths
but performs less well at larger strengths.
Together, these comparisons support the benefit of jointly
selecting where to intervene and which modality to enhance
based on the annotated functional roles.

\section{Additional Analysis of Token Role Assignment}
\label{app:role_analysis}

\textbf{Alternative Metrics for Token Role Assignment.}
To isolate the effect of the role-assignment metric, we replace
the Fisher--Rao score $v_t^m$ with an alternative score $g_t^m$,
while retaining the same role-assignment rule and downstream
advantage reweighting.
The Entropy variant uses the entropy of the within-modality
attention distribution:
\[
g_t^m
=
-\sum_{j\in\mathcal{I}_m}
p_{t,j}^m \log p_{t,j}^m.
\]
The KL variant measures the divergence of the current
within-modality attention distribution from the preceding one:
\[
g_t^m
=
D_{\mathrm{KL}}\!\left(p_t^m \,\|\, p_{t-1}^m\right)
=
\sum_{j\in\mathcal{I}_m}
p_{t,j}^m
\log\frac{p_{t,j}^m}{p_{t-1,j}^m}.
\]
For the textual modality, the two distributions are restricted
to their shared prefix and renormalized before computing KL
divergence, following the Fisher--Rao comparison.
Both variants use Eq.~(8) with $v_t^m$ replaced by $g_t^m$
and $\tau=0.05$; tokens satisfying neither condition remain neutral.

\textbf{Evaluation of Fisher--Rao Role Assignment.}
We evaluate how well different attention-based measurements recover
the human functional-role annotations introduced in
Appendix~\ref{app:hal_analysis}.
We compare Fisher--Rao distance with Attention, Cosine Similarity,
Entropy, and KL Divergence.
Attention uses modality-specific attention mass, while Entropy
measures the concentration of the within-modality attention
distribution at each step.
Cosine Similarity and KL Divergence compare attention
distributions between consecutive steps.
Fisher--Rao measures the angular distance between their
square-root representations, providing a symmetric characterization
of temporal redistribution within each modality.

For each measurement, we report three complementary statistics.
\emph{Coverage} denotes the proportion of human-annotated tokens
that receive an explicit visual or textual role assignment.
\emph{Accuracy} is computed over these non-neutral predictions,
with neutral assignments excluded from the denominator.
Because the annotated set contains unequal numbers of visually-
and text-oriented tokens, we additionally report
\emph{balanced accuracy}, computed as the mean of the per-class
recalls over the non-neutral subset.
Role assignments are obtained using the default
stability margin $\tau=0.05$.

\begin{table}[H]
\scriptsize
\centering
\caption{Agreement between attention-based role assignments and
human token-role annotations.}
\vspace{2pt}
\label{tab:role_agreement}
\setlength{\tabcolsep}{6pt}
\renewcommand{\arraystretch}{1.15}
\begin{tabular}{lccc}
\toprule
\textbf{Measurement}
& \textbf{Coverage (\%)}
& \textbf{Accuracy (\%)}
& \textbf{Balanced Acc. (\%)} \\
\midrule
Attention & 26.2 & 60.5 & 55.3 \\
Cosine Similarity & 18.1 & 65.0 & 69.0 \\
Entropy & 24.7 & 67.0 & 56.3 \\
KL Divergence & 28.2 & 71.5 & \textbf{73.5} \\
\rowcolor{ModelGreen}
\textbf{Fisher--Rao (Ours)}
& 26.3 & \textbf{74.0} & 72.8 \\
\bottomrule
\end{tabular}
\end{table}

As shown in Table~\ref{tab:role_agreement}, Fisher--Rao achieves
the highest overall agreement with the human annotations, reaching
74.0\% accuracy on its non-neutral predictions. Its balanced
accuracy (72.8\%) is also comparable to that of KL divergence
(73.5\%), indicating that the higher overall accuracy is not
obtained by strongly favoring one functional role. Together with
its comparable prediction coverage, these results support
Fisher--Rao as an effective empirical proxy for the functional
roles used by DyCo-RL.

\paragraph{Distribution of Assigned Token Roles.}
We analyze 1,000 rollouts randomly sampled from GRPO+DyCo-RL training on ThinkLite-hard-11K, using the training stability margin $\tau=0.05$. Table~\ref{tab:role_distribution} summarizes the resulting token-role distribution. Overall, 29.7\% of tokens are assigned visually-oriented or text-oriented roles and selected for role-conditioned advantage reweighting, while the remaining 70.3\% retain their original sequence-level advantages. Neutral assignments indicate insufficient separation between the visual and textual attention changes under the threshold $\tau$, rather than an absence of visual or textual processing.

\begin{table}[htbp]
\scriptsize
\centering
\caption{Distribution of token roles assigned by the Fisher--Rao criterion with $\tau=0.05$.}
\label{tab:role_distribution}
\setlength{\tabcolsep}{8pt}
\renewcommand{\arraystretch}{1.15}
\begin{tabular}{lc}
\toprule
Token role & Proportion (\%) \\
\midrule
Visually-oriented ($\mathcal{D}^{\mathrm{vis}}$) & 10.9 \\
Text-oriented ($\mathcal{D}^{\mathrm{txt}}$) & 18.8 \\
Neutral & 70.3 \\
\bottomrule
\end{tabular}
\end{table}

\textbf{Sensitivity to the Stability Margin $\tau$.}
The stability margin $\tau$ is introduced to prevent small and potentially ambiguous differences between visual and textual attention transitions from forcing a binary role assignment. A token is assigned a functional role only when one modality transition sufficiently dominates the other; otherwise, it is treated as neutral, excluded from the role-aware normalization set, and retains the original sequence-level advantage. 

Importantly, this neutral region also prevents tokens with no clear modality dominance, including potentially hybrid tokens that jointly depend on visual grounding and textual context, from being artificially pushed toward either modality. Thus, DyCo-RL does not impose a mutually exclusive visual/textual partition over all reasoning tokens; the binary role distinction is applied only to tokens exhibiting sufficiently confident modality-specific dynamics, while ambiguous cross-modal tokens remain unchanged. The sensitivity experiment therefore examines not only which value produces the highest score, but also whether DyCo-RL remains effective under different degrees of role-assignment conservativeness. We evaluate $\tau\in\{0,0.05,0.1\}$ on both Qwen2.5-VL-3B and Qwen2.5-VL-7B while keeping the remaining configuration fixed.

\begin{table}[H]
\scriptsize
\centering
\caption{Sensitivity analysis of the token role assignment stability margin $\tau$.}
\label{tab:tau_sensitivity}
\setlength{\tabcolsep}{4pt}
\renewcommand{\arraystretch}{1.15}
\begin{adjustbox}{max width=\linewidth}
\begin{tabular}{llcccccccc}
\toprule
\textbf{Model} & \textbf{$\tau$} & \textbf{WeMath} & \textbf{MathVision} & \textbf{MathVerse} & \textbf{MMBench} & \textbf{LogicVista} & \textbf{HallusionBench} & \textbf{MME} & \textbf{Avg.} \\
\midrule
Qwen2.5-VL-3B & GRPO & 23.0 & 21.4 & 34.4 & 72.8 & 39.8 & 60.8 & 79.9 & 47.4 \\

& 0 & 25.1 & 20.7 & 35.6 & 74.6 & \textbf{41.8} & 61.8 & 80.6 & 48.6 \\

\rowcolor{ModelGreen}
& \textbf{0.05} & \textbf{25.9} & 22.4 & \textbf{36.0} & \textbf{76.2} & 41.2 & \textbf{62.4} & \textbf{80.7} & \textbf{49.3} \\

& 0.1 & 25.2 & \textbf{22.7} & 35.2 & 75.8 & 40.7 & 61.5 & 80.1 & 48.7 \\
\midrule
Qwen2.5-VL-7B & GRPO & 38.7 & 26.6 & 47.4 & 78.0 & 46.1 & 69.2 & 83.3 & 55.6 \\
& 0 & 40.0 & 28.0 & 49.3 & 81.8 & 48.1 & 70.0 & 83.3 & 57.2 \\
\rowcolor{ModelGreen}
& \textbf{0.05} & \textbf{40.5} & \textbf{29.3} & \textbf{49.5} & \textbf{83.8} & \textbf{48.5} & \textbf{71.6} & \textbf{84.9} & \textbf{58.3} \\
& 0.1 & 40.3 & 27.3 & 49.2 & 82.9 & 48.1 & 69.2 & 84.6 & 57.4 \\
\bottomrule
\end{tabular}
\end{adjustbox}
\end{table}

Table~\ref{tab:tau_sensitivity} shows that DyCo-RL improves over GRPO under every tested value of $\tau$ on both model sizes. Its effectiveness therefore does not depend on a narrowly tuned threshold. Among the tested values, $\tau=0.05$ achieves the highest average performance for both Qwen2.5-VL-3B and Qwen2.5-VL-7B. This result suggests that a moderate margin provides an effective balance: it filters uncertain modality transitions while retaining tokens with sufficiently informative functional preferences. Notably, $\tau=0.05$ achieves higher average performance than $\tau=0$ on both model sizes, suggesting that forcing every token into a visual or textual role may not be necessary. Preserving a neutral region can help avoid assigning modality-specific roles to tokens whose multimodal function cannot be confidently reduced to a single dominant modality.

\section{Additional Ablation Studies}
\label{app:abl}

\paragraph{Ablation on Reweight Strength ($\alpha$).}

Table~\ref{tab:ablation_l} presents the ablation study on the reweight strength $\alpha$, which controls the intensity of modality-aware advantage adjustment during RL training. Overall, moderate reweighting strengths improve average visual reasoning performance over the unweighted baseline ($\alpha=0$). Specifically, $\alpha=0.2$ achieves the highest average performance, while $\alpha=0.1$ and $\alpha=0.3$ obtain the best results on several individual benchmarks. This suggests that moderate reweighting can effectively incorporate modality-aware guidance while preserving stable reasoning dynamics.

Conversely, when $\alpha$ is too small (e.g., $\alpha=0.1$), the modulation is relatively limited. At the other extreme, larger values ($\alpha \geq 0.4$) degrade average performance, with $\alpha=0.4$ already falling below the GRPO baseline. We hypothesize that overly aggressive reweighting may disrupt the model's intrinsic coordination and lead to less effective optimization. Overall, the results favor moderate reweighting strengths, while also revealing sensitivity to excessively large $\alpha$ values.

\begin{table*}[t]
\scriptsize
\centering
\caption{Ablation study on reweight strength $\alpha$. Bold formatting indicates the highest performance in each column.}
\label{tab:ablation_l}
\renewcommand{\arraystretch}{1.15}

\setlength{\tabcolsep}{4pt}
\begin{adjustbox}{max width=\textwidth}
\begin{tabular}{lcccccccc}
\toprule

\textbf{Method}
& \textbf{WeMath}
& \textbf{MathVision}
& \textbf{MathVerse}
& \textbf{MMBench}
& \textbf{LogicVista}
& \textbf{HallusionBench}
& \textbf{MME}
& \textbf{Avg.} \\

\midrule

GRPO Baseline ($\alpha=0$)
& 23.0
& 21.4
& 34.4
& 72.8
& 39.8
& 60.8
& 79.9
& 47.4 \\

$\alpha=0.1$
& 25.7
& 23.4
& 34.5
& 74.3
& 39.8
& \textbf{63.4}
& 80.1
& 48.7 \\

\rowcolor{ModelGreen}
\textbf{DyCo-RL ($\alpha=0.2$)}
& \textbf{25.9}
& 22.4
& 36.0
& \textbf{76.2}
& \textbf{41.2}
& 62.4
& 80.7
& \textbf{49.3} \\

$\alpha=0.3$
& 25.0
& \textbf{24.7}
& \textbf{36.4}
& 75.1
& 40.3
& 59.4
& \textbf{81.7}
& 48.9 \\

$\alpha=0.4$
& 23.1
& 19.4
& 32.4
& 74.9
& 37.4
& 59.9
& 79.0
& 46.6 \\

$\alpha=0.5$
& 21.0
& 17.8
& 31.5
& 71.1
& 35.3
& 58.6
& 78.2
& 44.8 \\
\bottomrule
\end{tabular}
\end{adjustbox}
\end{table*}

\paragraph{Advantage Reweighting vs.\ Reward Shaping.}
As shown in Table~\ref{tab:ablation_shaping}, we further contrast our token-level advantage reweighting with a straightforward trajectory-level reward shaping baseline. We aggregate the token-level alignment scores over the response as $\bar{s}=\frac{1}{T}\sum_{t=1}^{T}s_t$ and augment the accuracy reward:
\begin{equation}
R' = R_{\mathrm{acc}} + \beta \bar{s},
\end{equation}
where $\beta=0.2$ is the reward shaping coefficient. All other reward components and training settings remain unchanged.

While reward shaping improves the average performance over the vanilla GRPO baseline and performs favorably on some individual benchmarks, its overall performance remains below advantage reweighting. We hypothesize that this difference arises partly from their distinct interaction with GRPO optimization. Reward shaping modifies the trajectory-level reward before group normalization and consequently changes the resulting group-relative advantage estimates. Advantage reweighting, by contrast, introduces token-level alignment guidance after the baseline computation, providing more fine-grained modulation without directly altering the trajectory-level reward.

\begin{table*}[t]
\scriptsize
\centering
\caption{Comparison between Advantage Reweighting and Reward Shaping. Bold formatting indicates the highest performance in each column.}
\label{tab:ablation_shaping}

\renewcommand{\arraystretch}{1.15}
\setlength{\tabcolsep}{4pt}
\begin{adjustbox}{max width=\linewidth}
\begin{tabular}{lcccccccc}
\toprule

\textbf{Method}
& \textbf{WeMath}
& \textbf{MathVision}
& \textbf{MathVerse}
& \textbf{MMBench}
& \textbf{LogicVista}
& \textbf{HallusionBench}
& \textbf{MME}
& \textbf{Avg.} \\

\midrule

GRPO Baseline
& 23.0
& 21.4
& 34.4
& 72.8
& 39.8
& 60.8
& 79.9
& 47.4 \\

DyCo-RL (Reward Shaping)
& 24.4
& 19.7
& 34.5
& 75.5
& 37.6
& \textbf{62.4}
& 80.3
& 47.8 \\

\rowcolor{ModelGreen}
\textbf{DyCo-RL (Adv.\ Reweighting)}
& \textbf{25.9}
& \textbf{22.4}
& \textbf{36.0}
& \textbf{76.2}
& \textbf{41.2}
& \textbf{62.4}
& \textbf{80.7}
& \textbf{49.3} \\

\bottomrule
\end{tabular}
\end{adjustbox}
\end{table*}
\paragraph{Scaling the Number of Training Rollouts.}
The number of rollouts per prompt, $G$, controls the trade-off between training cost and performance. We evaluate DyCo-RL with $G \in \{4, 8, 16\}$ while holding all other hyperparameters constant. As shown in Table~\ref{tab:rollout}, its average score increases from 49.3 to 49.9 and 50.7, respectively. These results show that DyCo-RL benefits from larger rollout groups within the evaluated range, although they require additional computation for rollout generation and policy optimization. We adopt $G=4$ as the default configuration for resource efficiency.

\begin{table*}[h]
\scriptsize
\centering
\caption{Ablation Study on the Number of Training Rollouts ($G$).}
\label{tab:rollout}

\renewcommand{\arraystretch}{1.15}

\setlength{\tabcolsep}{4pt}
\begin{adjustbox}{max width=\linewidth}
\begin{tabular}{lcccccccc}
\toprule

\textbf{Method}
& \textbf{WeMath}
& \textbf{MathVision}
& \textbf{MathVerse}
& \textbf{MMBench}
& \textbf{LogicVista}
& \textbf{HallusionBench}
& \textbf{MME}
& \textbf{Avg.} \\

\midrule

\rowcolor{ModelGreen}
\textbf{DyCo-RL ($G=4$, Default)}
& 25.9
& 22.4
& \textbf{36.0}
& 76.2
& 41.2
& 62.4
& 80.7
& 49.3 \\

DyCo-RL ($G=8$)
& 27.1
& 23.0
& 35.3
& 77.6
& 41.4
& 62.7
& 82.0
& 49.9 \\

DyCo-RL ($G=16$)
& \textbf{28.0}
& \textbf{23.4}
& \textbf{36.0}
& \textbf{79.3}
& \textbf{42.3}
& \textbf{63.6}
& \textbf{82.5}
& \textbf{50.7} \\
\bottomrule
\end{tabular}
\end{adjustbox}
\end{table*}

\paragraph{Centered Advantage Reweighting.}
Our formulation applies the role-aware factor
$1+\alpha\operatorname{sgn}(\hat{A})w_t$
to non-neutral tokens.
Since $w_t$ averages to one over $\mathcal{D}$, this factor
introduces a common shift in addition to relative token-level
reweighting.
To examine whether the improvement persists after removing
this common component, we evaluate a centered variant:
\begin{equation}
\tilde{A}^{\mathrm{center}}_{t}
=
\begin{cases}
\big[1+\alpha\operatorname{sgn}(\hat{A})(w_t-1)\big]\hat{A},
& t\in\mathcal{D},\\
\hat{A},
& t\notin\mathcal{D}.
\end{cases}
\end{equation}
Because $\sum_{t\in\mathcal{D}}(w_t-1)=0$, the centered
variant preserves the mean token-level advantage within
each response, while retaining alignment-dependent differences
between tokens. Neutral tokens remain unchanged.
This preservation of the mean advantage does not imply
preservation of the policy-gradient norm.

\begin{table}[H]
\scriptsize
\centering
\caption{Comparison between centered and original advantage reweighting on GRPO.}
\label{tab:centered_reweighting}
\setlength{\tabcolsep}{4pt}
\renewcommand{\arraystretch}{1.15}

\begin{adjustbox}{max width=\linewidth}
\begin{tabular}{lcccccccc}
\toprule
\textbf{Method (Qwen2.5-VL-3B)}
& \textbf{WeMath}
& \textbf{MathVision}
& \textbf{MathVerse}
& \textbf{MMBench}
& \textbf{LogicVista}
& \textbf{HallusionBench}
& \textbf{MME}
& \textbf{Avg.} \\
\midrule

GRPO
& 23.0
& 21.4
& 34.4
& 72.8
& 39.8
& 60.8
& 79.9
& 47.4 \\

GRPO + DyCo-RL (centered)
& 25.3
& \textbf{23.4}
& 35.7
& \textbf{76.2}
& 39.1
& \textbf{62.4}
& \textbf{80.7}
& 49.0 \\

\rowcolor{ModelGreen}
\textbf{GRPO + DyCo-RL}
& \textbf{25.9}
& 22.4
& \textbf{36.0}
& \textbf{76.2}
& \textbf{41.2}
& \textbf{62.4}
& \textbf{80.7}
& \textbf{49.3} \\
\bottomrule
\end{tabular}
\end{adjustbox}
\end{table}

As shown in Table~\ref{tab:centered_reweighting}, the centered
variant also improves average performance over GRPO.
This result suggests that the improvement does not depend
solely on the common shift in token-level advantages and
supports the benefit of alignment-guided relative reweighting.

\section{Multi-Seed Evaluation}
\label{app:multiseed}

Reinforcement-learning optimization can vary across random seeds.
To examine whether the improvements of DyCo-RL are sensitive to
training randomness, we conduct additional experiments with three
training seeds, 42, 43, and 44, controlling training-data shuffling
and rollout sampling. The evaluation seed and decoding settings
are held fixed across runs. For each training seed, GRPO and
GRPO + DyCo-RL are trained on Qwen2.5-VL-3B and Qwen2.5-VL-7B
using otherwise identical training configurations.

\begin{table}[t]
\scriptsize
\vspace{-10pt}
\centering
\caption{Multi-seed evaluation. Each entry reports mean $\pm$ standard deviation over three matched random seeds.}
\label{tab:multiseed}
\setlength{\tabcolsep}{4pt}
\renewcommand{\arraystretch}{1.15}

\begin{adjustbox}{max width=\linewidth}
\begin{tabular}{lcccccccc}
\toprule
\textbf{Model} 
& \textbf{WeMath} 
& \textbf{MathVision} 
& \textbf{MathVerse} 
& \textbf{MMBench} 
& \textbf{LogicVista} 
& \textbf{HallusionBench} 
& \textbf{MME} 
& \textbf{Avg.} \\
\midrule

\rowcolor{gray!15}
\multicolumn{9}{c}{\textbf{Qwen2.5-VL-3B}} \\

GRPO
& 23.0\textsubscript{$\pm0.2$}
& 21.3\textsubscript{$\pm0.2$}
& 34.4\textsubscript{$\pm0.2$}
& 72.8\textsubscript{$\pm0.4$}
& 39.9\textsubscript{$\pm0.3$}
& 60.8\textsubscript{$\pm0.2$}
& 79.8\textsubscript{$\pm0.3$}
& 47.4\textsubscript{$\pm0.3$}
\\

GRPO + DyCo-RL
& 25.9\textsubscript{$\pm0.3$}
& 22.2\textsubscript{$\pm0.4$}
& 35.9\textsubscript{$\pm0.2$}
& 76.2\textsubscript{$\pm0.6$}
& 40.8\textsubscript{$\pm0.4$}
& 62.4\textsubscript{$\pm0.2$}
& 80.8\textsubscript{$\pm0.3$}
& 49.2\textsubscript{$\pm0.2$}
\\

\midrule

\rowcolor{gray!15}
\multicolumn{9}{c}{\textbf{Qwen2.5-VL-7B}} \\

GRPO
& 38.6\textsubscript{$\pm0.2$}
& 26.6\textsubscript{$\pm0.4$}
& 47.6\textsubscript{$\pm0.2$}
& 78.4\textsubscript{$\pm0.5$}
& 46.4\textsubscript{$\pm0.5$}
& 69.2\textsubscript{$\pm0.3$}
& 83.6\textsubscript{$\pm0.4$}
& 55.8\textsubscript{$\pm0.3$}
\\

GRPO + DyCo-RL
& 40.3\textsubscript{$\pm0.3$}
& 29.0\textsubscript{$\pm0.4$}
& 49.5\textsubscript{$\pm0.3$}
& 83.5\textsubscript{$\pm0.5$}
& 48.1\textsubscript{$\pm0.5$}
& 71.7\textsubscript{$\pm0.3$}
& 84.8\textsubscript{$\pm0.4$}
& 58.1\textsubscript{$\pm0.3$}
\\

\bottomrule
\end{tabular}
\end{adjustbox}
\vspace{-10pt}
\end{table}

Table~\ref{tab:multiseed} reports the mean and standard deviation across the three runs. DyCo-RL consistently improves over GRPO across all evaluated benchmarks and both model scales. In terms of average performance, DyCo-RL improves the score from 47.4 to 49.2 on Qwen2.5-VL-3B and from 55.8 to 58.1 on Qwen2.5-VL-7B. The standard deviations remain modest across repeated runs, suggesting that the observed gains are consistently preserved under the training variability examined here rather than being attributable to a particular random seed.
\section{Analysis of Computational Overhead}
\label{app:overhead}

As detailed in Table~\ref{tab:overhead}, DyCo-RL reduces
training throughput on Qwen2.5-VL-3B from 0.447 to 0.327 samples/sec, a decrease of 26.8\%. 

The overhead primarily arises from extracting and processing
attention weights during training rollouts, including explicit
attention materialization and modality-wise aggregation.
Fisher--Rao distance computation and token role assignment
introduce additional processing.
Attention statistics are collected during autoregressive
rollout generation, without a separate full-model forward
pass after the rollout solely for attention extraction. 
\begin{table}[H]
\scriptsize
\centering
\caption{Analysis of training throughput and computational overhead.}
\label{tab:overhead}

\setlength{\tabcolsep}{8pt} 
\renewcommand{\arraystretch}{1.15}

\begin{adjustbox}{max width=\linewidth}
\begin{tabular}{lcc}
\toprule
\textbf{Method (Qwen2.5-VL-3B)} & \textbf{Throughput (samples/sec)} & \textbf{$\Delta$ Throughput} \\
\midrule
GRPO  & 0.447 & -- \\
\addlinespace[0.3em] 
\textbf{GRPO + DyCo-RL} & \textbf{0.327} & \textbf{-26.8\%} \\
\bottomrule
\end{tabular}
\end{adjustbox}
\end{table}

These additional operations are required only during training.
At inference time, the trained model can use the standard
backbone inference pipeline without attention-statistics
extraction, role assignment, or advantage reweighting.
DyCo-RL therefore introduces no additional trainable parameters
or method-specific inference computation, while incurring
the training overhead reported above.

\section{Analysis of Training Dynamics}
\label{app:dyn}
To elucidate how DyCo-RL reshapes the optimization landscape, we compare its training trajectory against the standard GRPO baseline with respect to the accuracy reward. For visual clarity, the plotted curves are smoothed using a rolling mean with a window size of 10. 

\begin{figure}[H]
    \centering
    \includegraphics[width=0.5\linewidth]{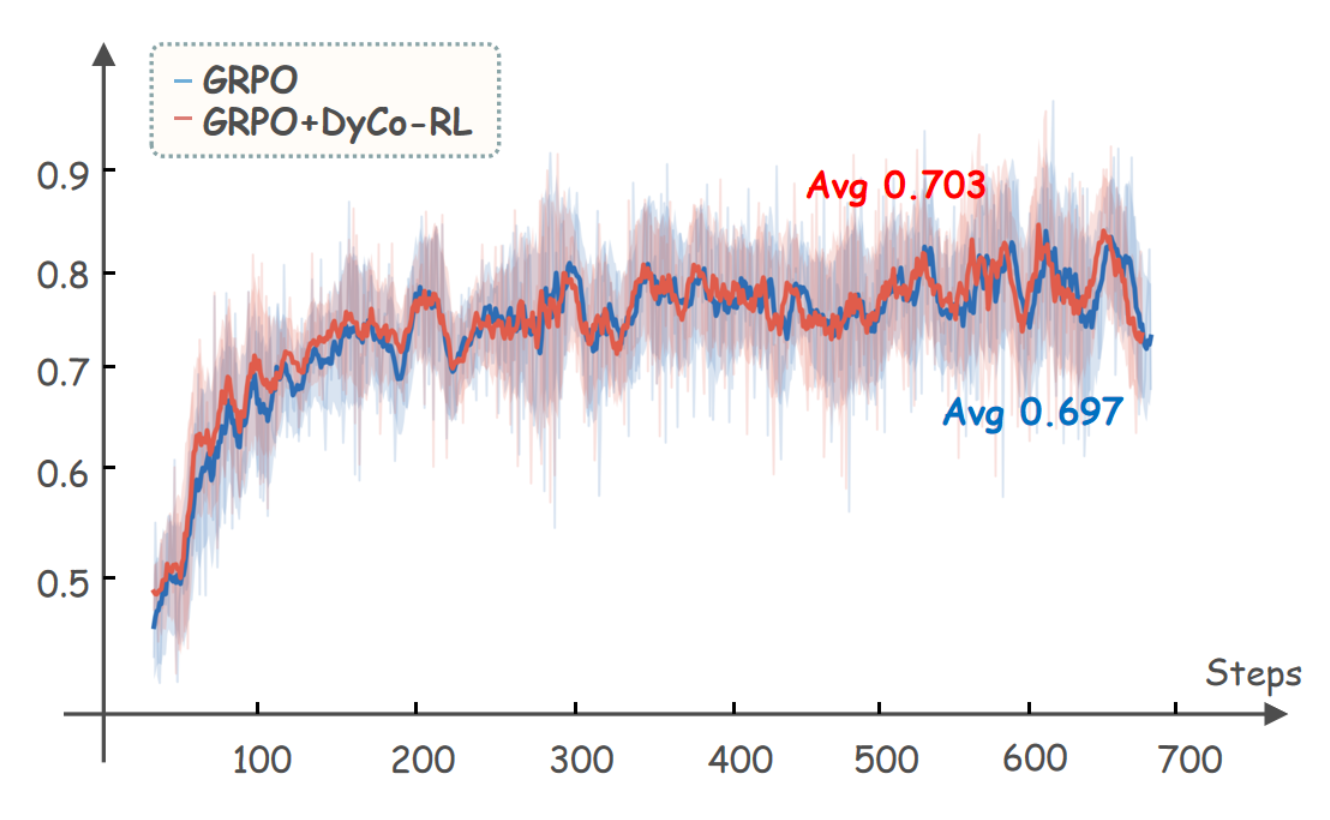}
    \caption{Training accuracy reward of GRPO and GRPO + DyCo-RL. Curves are smoothed using a rolling mean with a window size of 10.}
    \label{fig:training_dynamics}
\end{figure}

As illustrated in Fig.~\ref{fig:training_dynamics}, both methods
show rapid initial increases in accuracy reward, followed by
fluctuations around similar levels. DyCo-RL achieves a slightly
higher average reward over the training period (0.703 versus
0.697), although the curves overlap and cross at multiple stages.
This comparison suggests a modest overall benefit of role-aware
advantage reweighting for training reward optimization.

\section{Extended Generalization Analysis}
\label{app:gen}

A critical question regarding token-level advantage reweighting is whether the learned alignment signal overfits to the visual-mathematical distribution seen during training. To assess its out-of-domain robustness, we further evaluate DyCo-RL on {A-OKVQA~\citep{10.1007/978-3-031-20074-8_9}, RealWorldQA~\citep{realworldqa2024}, MMStar~\citep{chen2024we}, SEED-Bench~\citep{li2023seed}, ChartQA~\citep{masry-etal-2022-chartqa}, and MMMU~\citep{yue2023mmmu}, covering commonsense visual understanding, real-world reasoning, general multimodal understanding, chart reasoning, and multidisciplinary expert-level reasoning.
\vspace{-7pt}
\begin{table}[h]
\scriptsize
\centering
\vspace{ 5pt}
\caption{Out-of-domain generalization results. Bold formatting indicates the higher performance on each benchmark.}
\label{tab}

\setlength{\tabcolsep}{4pt}
\renewcommand{\arraystretch}{1.15}

\begin{adjustbox}{max width=\linewidth}
\begin{tabular}{lcccccc}
\toprule

\textbf{Method (Qwen2.5-VL-3B)}
& \textbf{A-OKVQA}
& \textbf{RealWorldQA}
& \textbf{MMStar}
& \textbf{SEED-Bench (IMG)}
& \textbf{ChartQA (Test)}
& \textbf{MMMU (Val)} \\
\midrule

GRPO 
& 82.3
& 47.0
& 51.2
& 68.2
& 81.0
& 42.2 \\

\addlinespace[0.5em]

\rowcolor{ModelGreen}
\textbf{GRPO + DyCo-RL}
& \textbf{82.7}
& \textbf{49.5}
& \textbf{53.2}
& \textbf{69.0}
& \textbf{83.2}
& \textbf{45.2} \\

\bottomrule
\end{tabular}
\end{adjustbox}

\end{table}

As shown in Table~\ref{tab}, DyCo-RL consistently outperforms GRPO across all six  benchmarks, demonstrating that the learned alignment strategy generalizes beyond the visual-mathematical training distribution to diverse multimodal reasoning and understanding tasks. These consistent gains across different domains and task formats further support the generalizability between domains of the proposed role-aware reweighting.

\section{Prompt Template}
\label{app:prompt}
We adopt a GRPO-style reasoning format that requires the model to generate intermediate reasoning within <think> tags and final responses within <answer> tags:
\begin{tcolorbox}[colback=gray!10, colframe=gray!50, title=Prompt Template]
\textbf{SYSTEM:} You are a helpful assistant.

\vspace{0.5em}
\textbf{USER:}A conversation between User and Assistant. The user asks a question, and the Assistant solves it.
The assistant first thinks about the reasoning process in the mind and then provides the user with the answer.
The reasoning process and answer are enclosed within <think> </think> and <answer> </answer> tags, respectively.\\
\textbf{Question}: \{question\}
\end{tcolorbox}
The template above shows the textual component of the prompt, images are supplied through the model's standard multimodal processor and chat template.

\section{More Visualization Results}
\label{app:vis}

We provide additional qualitative examples in Fig.~\ref{fig:vis11}--\ref{fig:vis3}, comparing the standard GRPO baseline against our DyCo-RL augmented version on the Qwen2.5-VL-3B architecture. Red text marks erroneous reasoning spans in the baseline responses, while green text marks correct reasoning spans in our model's responses.
The selected examples illustrate how DyCo-RL can interleave visual evidence extraction and text-based reasoning, complementing the quantitative analyses in the main paper.
\newpage

\begin{figure*}
    \centering
    \includegraphics[width=1\linewidth]{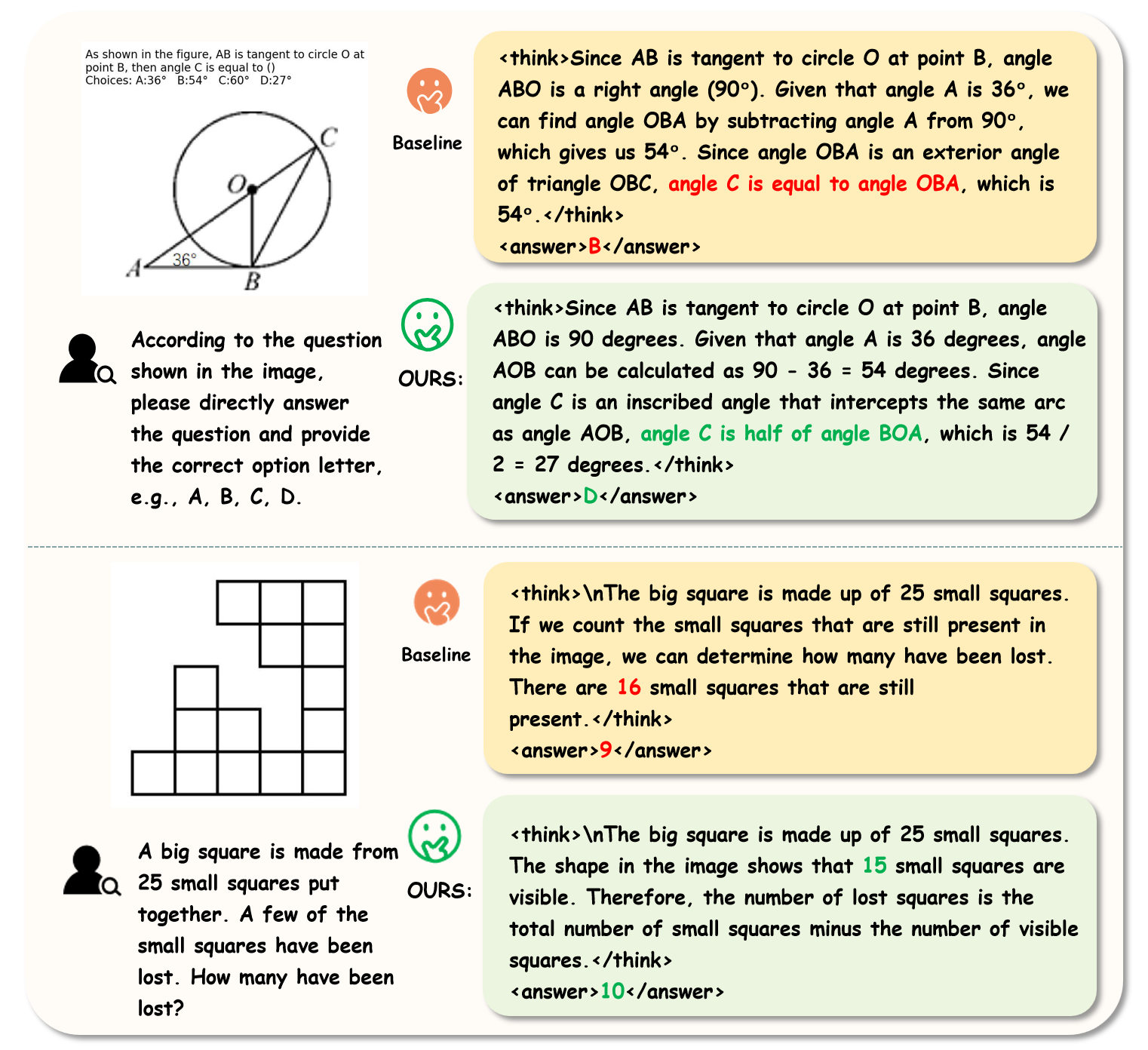}
    \caption{Visualization Cases on Mathematical and  Logical Reasoning.}
    \label{fig:vis11}
\end{figure*}

\begin{figure*}
    \centering
    \includegraphics[width=1\linewidth]{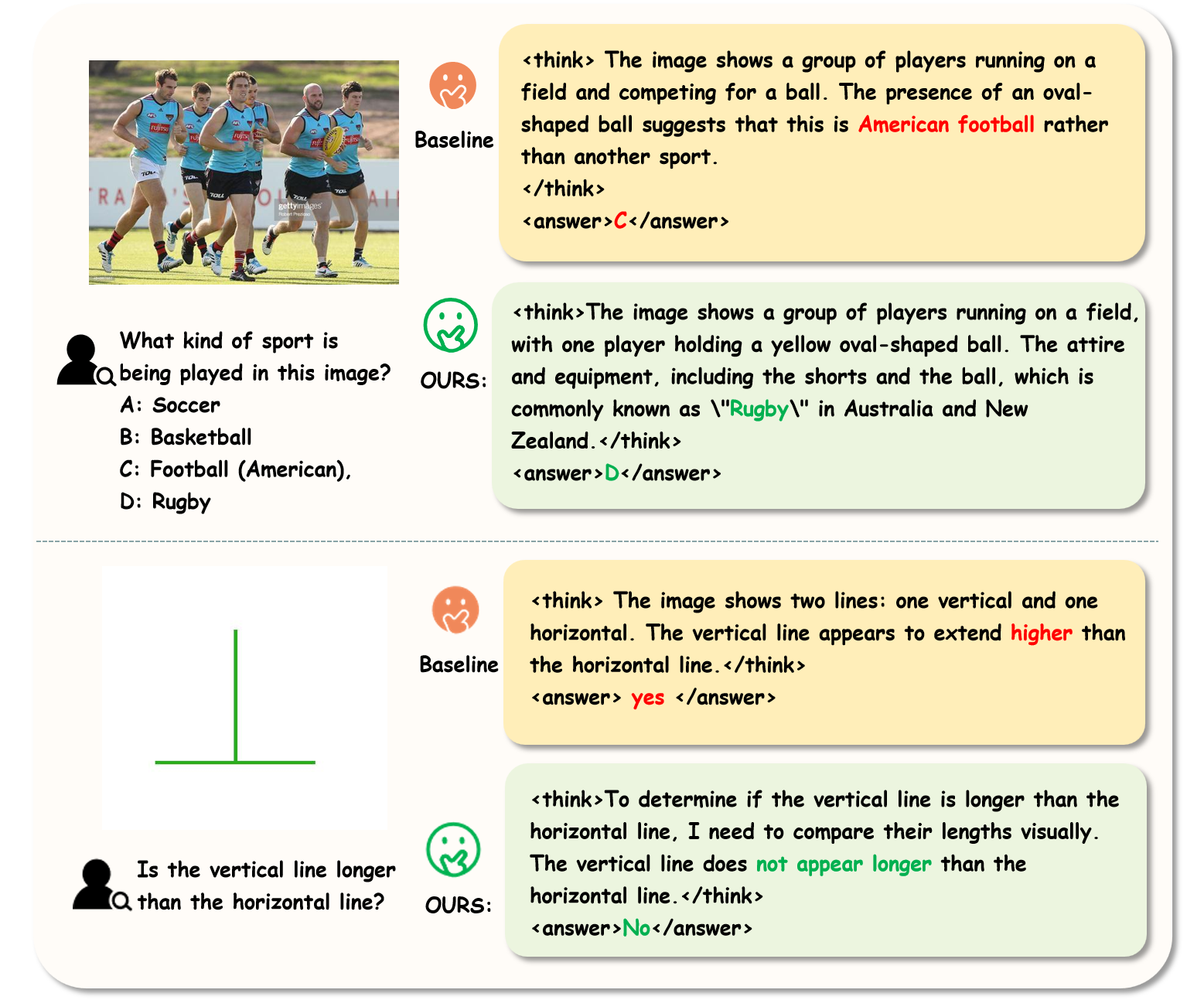}
    \caption{Visualization Cases on General Task and Hallucination Reasoning.}
    \label{fig:vis3}
\end{figure*}


\end{document}